\documentclass[12pt]{article}
 \usepackage{natbib}
\usepackage{graphicx,psfrag,epsf}
\usepackage{enumerate}
\usepackage{times,epsfig,graphics,amssymb,latexsym,amsmath,amsthm,fullpage, setspace, algorithm, algorithmic}
\usepackage{array,hyperref,url}
\usepackage[usenames]{color}
\usepackage{enumitem}
\usepackage{amsmath}
\usepackage{multirow}
\usepackage[titletoc]{appendix}
\usepackage{caption}
\usepackage{threeparttable}
\newlist{steps}{enumerate}{1}
\setlist[steps, 1]{label = Step \arabic*:}

\newtheorem{theorem}{Theorem}

\newtheorem{lemma}{Lemma}

\allowdisplaybreaks[4]

\def\argmin{\mathop{\rm argmin}}

\def\Var{\mathop{\rm Var}}

\begin{document}

\pagenumbering{arabic}

\title{\Large \bf Efficient Learning of Quadratic Variance Function Directed Acyclic Graphs via Topological Layers}
\author{Wei Zhou\footnote{School of Data Science, City University of Hong Kong} \footnote{Wang Yanan Institute for Studies in Economics, Department of Statistics, School of Economics, Fujian Key Lab of Statistics and MOE Key Lab of Econometrics, Xiamen University} , Xin He\footnote{School of Statistics and Management, Shanghai University of Finance and Economics} , Wei Zhong$^{\dagger}$, Junhui Wang$^{\ast}$
}
\date{}

\maketitle

\setcounter{page}{1}

\begin{abstract}
Directed acyclic graph (DAG) models are widely used to represent casual relationships  among random variables in many application domains.  This paper studies a special class of non-Gaussian DAG models, where the conditional variance of each node given its parents is a quadratic function of its conditional mean.  Such a class of non-Gaussian DAG models are fairly flexible and admit many popular distributions as special cases,  including Poisson, Binomial, Geometric, Exponential, and Gamma. To facilitate learning, we introduce a novel concept of topological layers, and  develop an efficient DAG learning algorithm. It first reconstructs the  topological layers in a hierarchical fashion and then recoveries the directed edges between nodes in different layers, which requires much less computational cost than most existing algorithms in literature. Its advantage is also demonstrated in a number of simulated examples, as well as its applications to two real-life datasets, including an NBA player statistics data and a cosmetic sales data collected by Alibaba.
\end{abstract}
\bigskip
{\bf Key Words and Phrases:} Causality, quadratic variance function,  non-Gaussian DAG, structural equation model (SEM)

\onehalfspacing

\section{Introduction}

Directed acyclic graph (DAG) model plays a crucial role in causal inference, which is widely used to represent casual relationships among random variables, and has a wide range of applications  such as genetics, finance
and machine learning  \citep{sach, sanford, koller}. Yet, learning a DAG model from observed data is challenging both methodologically and computationally, largely due to the identifiability issue and the required acyclicity.

Most of the earlier DAG learning approaches in literature ignore the identifiability issue and mainly focus on recovering the Markov equivalent class \citep{Spirtes2000} of the DAG model. For example, the search-and-score algorithm \citep{chickering2003, Nandy, Zheng, Zhu} maximized the regularized likelihood for a DAG model with the best score, and the constrained-based method \citep{Kalisch, Spirtes2000,Tsama} first conducted local conditional independence tests to learn the skeleton of the DAG model, and then determined the edge directions based on acyclicity, v-structures, and other available structures. These methods are able to recover the Markov equivalent class of the DAG model under some assumptions, such as the Markov property and the faithfulness condition. To fully identify DAG models, a number of learning algorithms have been developed under various assumptions on the underlying probability distribution of the DAG model, often represented as a structural equation model (SEM). Particularly, \cite{buhlmann} established the identifiability of linear Gaussian SEM models with equal error variances, and various algorithms have been developed accordingly \citep{buhlmann,chen,Yuan2019}. Following this line, \cite{shimizu2006, shimizu2011} and \cite{WangYS2020} established the identifiability of linear non-Gaussian SEM model  and developed the corresponding learning algorithms, and \cite{aosbuhl} and \cite{peters} established the identifiability of nonparametric SEM model with additive noise assumption.

More recently,  \cite{Park20191} and \cite{Park20192} studied a general class of non-Gaussian DAG models, denoted as QVF-DAGs, which require that the conditional variance of each node given its parents is a quadratic function of its conditional mean. This assumption provides a natural criterion for determining the causal ordering of nodes without making additional distributional assumptions, and contains many non-Gaussian distributions as its special cases. An over-dispersion scoring (ODS; \cite{Park20191}) algorithm is also developed for learning QVF-DAG, which first estimates the moral graph of the QVF-DAG model and learns the causal ordering of all nodes sequentially, and then determines the parents of each node through some sparse regression models over the nodes which are causally ahead of it. Yet the computational cost of the ODS algorithm is usually expensive even for learning a medium-size QVF-DAG model.

In this article,  we propose a computationally efficient learning algorithm for QVF-DAGs based on a novel concept of topological layers. The idea is very intuitive and any DAG model can be reorganized into an equivalent topological structure with multiple layers, which automatically ensures acyclicity as a node can only have children and offsprings in its lower layers.  More importantly, QVF-DAGs can be reconstructed via topological layers  based on the proposed ratio-based criterion, which can be used to hierarchically determine the membership of each layer. Once the layers are determined, parents of each node can be recovered by applying some sparse regression methods over all nodes in its upper layers.

Compared with the ODS algorithm in \cite{Park20191}, the proposed learning algorithm has a number of advantages. First, the topological layers in the proposed algorithm are unique for any given QVF-DAG model, whereas the ODS algorithm needs to estimate the indeterministic causal ordering of nodes. Second, the computational cost of the proposed algorithm is much less than that of the ODS algorithm, especially in some wide yet shallow QVF-DAG models, such as a hub graph, which has attracted tremendous interest in network analysis as it is a main building block for many network architectures. More precisely, to learn a hub graph with $p$ nodes and $n$ samples as in Figure \ref{fig:hub},
the computational complexity of our proposed algorithm is of order $O(np)$, which is much more efficient than the ODS algorithm, whose computational complexity is of order $O(np^2)$ \citep{Park20191}.

The rest of this paper is organized as follows. Section 2 introduces QVF-DAGs and the concept of topological layers, and the reconstruction criterion of QVF-DAGs based on topological layers.  Section \ref{sec3} provides the details of the proposed learning algorithm for QVF-DAGs. Numerical experiments on simulated examples are conducted in Section \ref{sec4} to demonstrate the advantages of the proposed algorithm compared with some existing competitors. Section \ref{sec5} applies the proposed algorithm to analyze two real-life datasets, including an NBA player statistics data and a cosmetic sales data from Alibaba company. A summary is given in Section \ref{sec:sum}, and Appendix is devoted to some computational details and technical proofs.

\section{QVF-DAG and topological layers}
\label{sec2}

\subsection{QVF-DAG}

DAG models are widely used to encode joint distribution of a random vector $(X_1,...,X_p)$.
Precisely, let ${\cal G}=({\cal V}, {\cal E})$ denote a DAG, where  ${\cal V}=\{1,...,p\}$ represents a set of nodes associated with variables $X=(X_j)_{j \in {\cal V}}$,  and ${\cal E} \subset {\cal V}\times {\cal V}$ denote a set of directed edges without directed cycles.  A directed edge from node $k$ to node $j$ is denoted as $k \rightarrow j$, and then node $k$  is a parent of node $j$, and the set of node $j$'s parents is denoted as $\mbox{pa}_j$.
Let $X_{\text{pa}_j}:=\{ X_k: k \in \text{pa}_j \subset {\cal V}  \}$ and
$X_{\cal S} := \{ X_k:k \in {\cal S} \subset {\cal V}  \}$ for any subset ${\cal S}$ of ${\cal V}$.
We assume that the joint distribution $P(X)$ satisfies the Markov property with respect to ${\cal G}$, and thus it allows for the following factorization,
$$
P(X)=\prod_{j \in {\cal V}} P(X_j| X_{\text{pa}_j}),
$$
where $P(X_j| X_{\text{pa}_j})$ denotes the conditional distribution of $X_j$ given its parents $X_{\text{pa}_j}$.

In this paper, we focus on the non-Gaussian DAG models with the quadratic variance function (QVF) property \citep{Park20191}. Specifically, we assume that $P(X_j| X_{\text{pa}_j})$  satisfies the QVF property that there exist some constants $\beta_{j1}$ and $\beta_{j2}$ such that
\begin{align}\label{eqn:variance}
\Var(X_j| X_{\text{pa}_j}) = \beta_{j1} E [X_j| X_{\text{pa}_j}]  + \beta_{j2} \big (
E [X_j| X_{\text{pa}_j}]  \big )^2.
\end{align}
In literature, the natural exponential family with the QVF property has been  extensively studied
\citep{Morris, Brown}, which further assumes that $P(X_j| X_{\text{pa}_j})$ belongs to the exponential family,
\begin{eqnarray}\label{conditionaldis}
P(X_j| X_{\text{pa}_j}) = \exp \Big( \theta_{j} X_j +\sum_{k \in \text{pa}_j } \theta_{jk} X_k X_j - B_j(X_j) - A_j \Big(  \theta_{j} + \sum_{k \in \text{pa}_j}\theta_{jk} X_k \Big ) \Big),
\end{eqnarray}
where $A_j(\cdot)$ is a log-partition function, $B_j(\cdot)$ is determined by a given distribution in the exponential family, and the parameter $\theta_{jk} \in {\cal R}$ represents the effect from node $k$ to  node $j$. As pointed out in \cite{Park20191}, the QVF-DAG model is identifiable and
many popular non-Gaussian distributions satisfy both assumptions \eqref{eqn:variance} and \eqref{conditionaldis}, including Poisson, Binomial, Geometric, Exponential, Gamma, and so on.

\subsection{Topological layers}\label{sec:layer}

We introduce a novel concept of topological layers, to reformulate a QVF-DAG model into an equivalent topological structure with multiple layers. Particularly, all root nodes in the QVF-DAG model are assigned to the top layer, and other nodes are assigned to different layers according to their longest distances to a root node. It is also important to point out that the induced topological layers are unique for the given QVF-DAG model, and the parents of each node must belong to its upper layers, thus automatically assuring acyclicity. 

Suppose there are a total of $T$ layers, where $T$ denotes the longest possible distance from some node in the QVF-DAG model to a root node. Let ${\cal A}_0$ denote the set of root nodes and isolated nodes in the top layer, ${\cal A}_t$ denote the set of nodes in the  $(t+1)$-th topological layer, and ${\cal S}_t=\cup_{d=0}^{t-1}{\cal A}_d$ denote all the nodes in the layers above the $(t+1)$-th layer. For any node $k \in {\cal A}_t$, its parent nodes are denoted as $\text{pa}_k$, and thus $\text{pa}_k \subseteq {\cal S}_{t}$.

\begin{figure}[!h]
	\caption{A toy QVF-DAG model in the left panel, and its equivalent topological structure with three layers in the right panel.}\label{fig:0}
	\centering
	\includegraphics[width=0.8\textwidth]{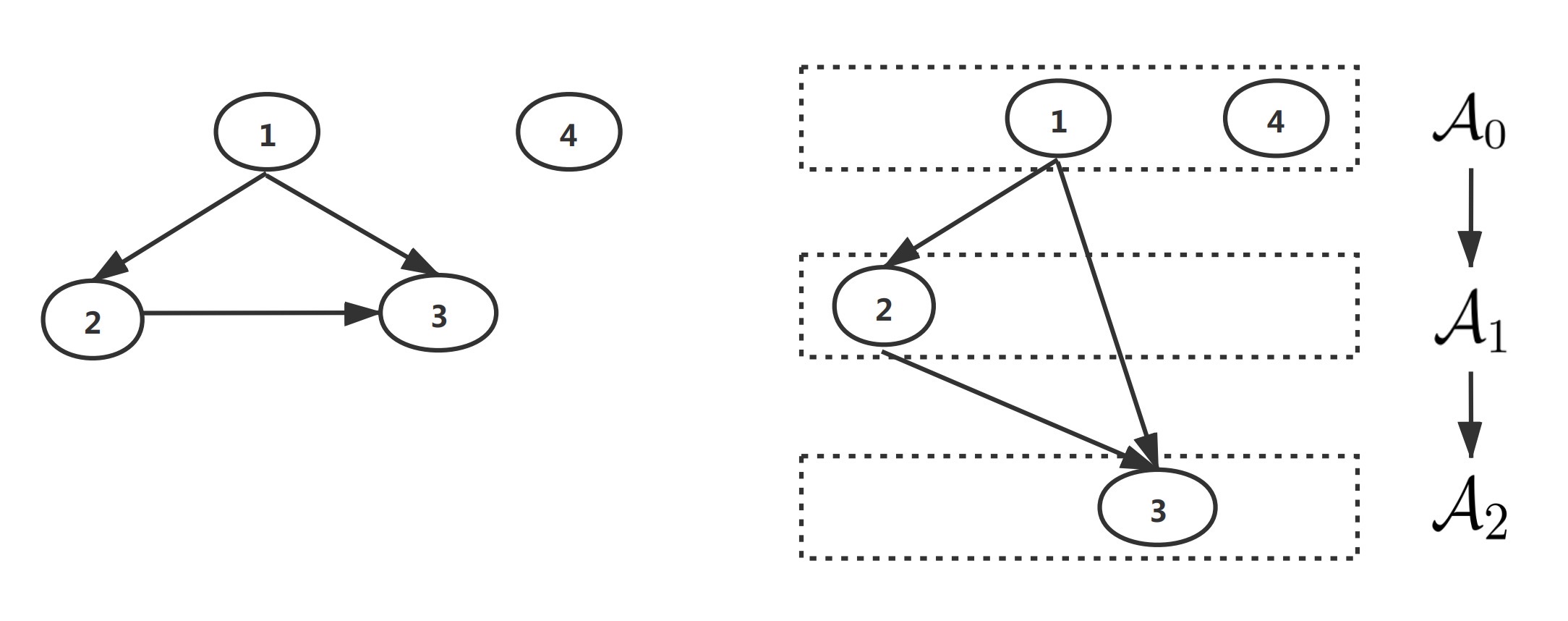}
\end{figure}

Figure \ref{fig:0} provides a toy QVF-DAG model with three topological layers. Note that node $1$  is a root node and node $4$ is an isolated node, and thus both nodes belong to ${\cal A}_0$.
Node $2$ belongs to ${\cal A}_1$ as its longest distance to the root node is 1, but node $3$ belongs to ${\cal A}_2$ instead of ${\cal A}_1$ since its longest path to the root node is $1 \rightarrow 2 \rightarrow 3$. Note that the  topological layers is closely related with the key ideas of the depth-first search (DFS) and breadth-first search (BFS) algorithms \citep{cormen2009introduction}. Specifically, the number of topological layer for a DAG is defined by the longest distance to a root node, which is similar to the DFS algorithm exploring as far as possible starting from a root node.  The idea that nodes with the same distance to a root node are in the same layer is similar to that of the BFS algorithm, which explores all of the neighbor nodes of the current depth before moving to the next depth level.

\subsection{Reconstruction of topological layers} \label{sec:identify}

Let $\text{nd}_j$ denote all the non-descendant nodes of node $j$ excluding itself, then the topological layers of a QVF-DAG model can be reconstructed under some mild technical conditions.

\noindent {\bf Condition 1:}  For any node $j \in {\cal V}$ and any set ${\cal S}$ satisfying that $\text{pa}_j \nsubseteq {\cal S} \subset \text{nd}_j$, we have
 $E\big[ \omega^2_j({\cal S})\Var(E[  X_j| X_{\text{pa}_j}  ]| X_{\cal S})\big ] >0$, and $\omega_j({\cal S})=( \beta_{j1}+\beta_{j2}E[  X_j| X_{\cal S} ]  )^{-1}$ exists.

Condition 1 is quite general and can be verified for many popular distributions, including Poisson, Binomial, Exponential, Gamma, Geometric, Negative Binomial and so on. The first part of Condition 1 requires all the parents of node $j$ should contribute to its variability, which is more general than that  in \cite{Park20191} assuming $\Var(E[ X_j| X_{\text{pa}_j}]| X_{\cal S} )>0$ for all $X_{\cal S}$, and can reduce to the identifiability condition in \cite{Park20192} when the underlying distribution is indeed Poisson. For illustration, we consider a toy example with $X_1 \sim \text{Poisson}(\lambda)$, $X_2|X_1\sim \text{Poisson}(\lambda+X_1)$ and $X_3|X_1,X_2\sim \text{Poisson}(\lambda+X_2 {1}_{  \{X_1\neq 1\}}  )$, where $1_{\{\cdot\}}$ is an indicator function. It can be verified that $\Var(E[X_3|X_1,X_2]|X_1=1 )=0$ and thus the identifiability condition in \cite{Park20191} is violated, but this toy example still satisfies the first part of Condition 1. The second part of Condition 1 requires that  $\beta_{j1}$ and $\beta_{j2}$ should satisfy  $\beta_{j1}+\beta_{j2}E[X_j| X_{\cal S}] \neq 0$, which ensures that $\omega_j({\cal{S}}) = ( \beta_{j1}+\beta_{j2} E[  X_j | X_{\cal S} ] )^{-1}$ is well-defined.
Moreover, $\beta_{j2}>-1$ is required to rule out some distributions, including Bernoulli and multinomial distributions, which are known to be unidentifiable in literature \citep{Heckerman1}. 

\begin{lemma}\label{lemma1}
Suppose  that  $X \in {\cal R}^p$ is generated by a QVF-DAG model and Condition 1 is satisfied. For any node $j \in {\cal V}$ and any set ${\cal S}  \subseteq \text{nd}_j$, we have
$$
E \big [\Var(\omega_j({\cal S}) X_j| X_{\cal S}) \big ] \geq    E[\omega_j({\cal S}) X_j ],
$$
provided that $\beta_{j2}>-1$.  The equality holds if and only if   $\text{pa}_j \subseteq  {\cal S}$.
\end{lemma}

Lemma \ref{lemma1} provides a crucial criterion to reconstruct the topological layers of a QVF-DAG model in a top-down fashion. Specifically, if $\text{pa}_j \subseteq  {\cal S}$,  the conditional ratio
\begin{align}
 {\cal R}(j, {\cal S}) := \frac{   E \big [\Var(\omega_j({\cal S}) X_j| X_{\cal S}) \big ]     }{  E[\omega_j({\cal S}) X_j  ]   }
\end{align}
should be exactly 1. Motivated by this fact, Theorem \ref{thm1} shows that the topological layers $\{{\cal A}_t\}_{t=0}^{T-1}$ of a QVF-DAG model, defined in Section \ref{sec:layer}, can be exactly reconstructed.

\begin{theorem}\label{thm1}
Suppose that  all the conditions of Lemma \ref{lemma1} are satisfied and   ${\cal A}_0,\dots,{\cal A}_{t-1}$ have been identified with ${\cal S}_0 = \emptyset$ and ${\cal S}_t=\cup_{d=0}^{t-1}{\cal A}_d$. It then  holds true that 
		\begin{align}
			{\cal R}(j, {\cal S}_t)
			\begin{cases}\label{eqn:motivation2}
				= 1,       &\quad \text{for any} \ j\in {\cal A}_t;\\
				\neq 1, & \quad \text{for any} \ j \in {\cal V}\backslash  \{ {\cal S}_{t} \cup {\cal A}_t \},
			\end{cases}
		\end{align}
for $t=0,...,T-1$,	and thus  the topological layers  can be exactly reconstructed.
\end{theorem}

Theorem \ref{thm1} provides a constructive result for the reconstruction of a general class of non-Gaussian DAG models with the QVF property. Its proof follows from Lemma \ref{lemma1}, the criterion in \eqref{eqn:motivation2}, and the assumption that a node's parents should all contribute to its variability. Particularly, for any root or isolated node $j \in {\cal V}$ with $\text{pa}_j=\emptyset$, Theorem \ref{thm1} shows  that  the topological layer ${\cal A}_0$  can be exactly reconstructed by the fact that
$$
{\cal R}(j, \emptyset)  = \frac{   E \big [\Var(\omega_j({\emptyset}) X_j) \big ]     }{  E[\omega_j({\emptyset}) X_j  ]   }
= \frac{  \Var( X_j )     }{  (\beta_{j1}+ \beta_{j2} E[X_j]   ) E[    X_j  ]   } = 1,
$$
for any root or isolated node $j \in {\cal V}$, and $ {\cal R}(j, {\emptyset} ) \neq 1$ otherwise. Further, if the longest distance of a node $j$ to a root node is $t \ge 1$, then $j \in {\cal A}_t$ by definition, and Theorem \ref{thm1} guarantees that ${\cal R}(j, {\cal S}_t)  = 1$ and ${\cal R}(l, {\cal S}_t) \neq 1$ for any node $l$ contained in lower layers.

\section{Proposed algorithm}\label{sec3}

In this section, we illustrate the proposed algorithm with natural exponential family assumption in \eqref{conditionaldis}, but the algorithm can be adapted to learn a general QVF-DAG model as well. Motivated by Lemma \ref{lemma1} and Theorem \ref{thm1}, learning a QVF-DAG model from the observed data can be decomposed into a two-step procedure, where the topological layers can be reconstructed in a top-down fashion at the first step, and then the directed edges can be recovered by using sparse regression models in a parallel fashion.

\subsection{Two-step learning algorithm}

Given a training sample $X^n=(  X_i^n )_{i=1}^n$ with  $X_i^n=(X_{i,1}^{n},...,X_{i,p}^{n})^T$, we first attempt to estimate the top layer ${\cal A}_0$. Specifically, for each node $j \in {\cal V}$,   we compute the estimated  unconditional ratio,
\begin{eqnarray}\label{root}
\widehat{\cal R}(j, \emptyset)=\frac{  \widehat{\Var} ( X_j  )     }{  (\beta_{j1}+ \beta_{j2} \widehat{E}[ X_j]   ) \widehat{E}[   X_j  ]   }  ,
\end{eqnarray}
where $\widehat{\Var} (    X_j     ) = \widehat{E}[X_j^2]- (\widehat{E}[    X_j  ] )^2,
\widehat{E}[    X_j  ] =\frac{1}{n} \sum_{i=1}^n X_{i,j}^{n} $ and $\widehat{E}[    X^2_j  ] =\frac{1}{n} \sum_{i=1}^n ( X_{i,j}^{n} )^2 $. By Theorem \ref{thm1}, ${\cal A}_0$ can be estimated as
$\widehat{\cal A}_{0}=\big \{j, \ |\widehat{\cal R}(j, \emptyset)-1 | \leq \epsilon_0 \big \}$ for  some small constant $\epsilon_0>0$.

Suppose that the topological layers $\widehat{\cal A}_0,\dots, \widehat{\cal A}_{t-1}$ have been estimated and $\widehat{\cal S}_t=\cup_{d=0}^{t-1} \widehat{\cal A}_d$, we now proceed to estimate ${\cal A}_t$. For each node  $ j\in {\cal V}\backslash \widehat{\cal S}_t,$ we compute the estimated conditional ratio,
\begin{align}\label{eqn:ratio}
\widehat{\cal R}(j, \widehat{\cal S}_t )=\frac{  \widehat{ E} \big [ \widehat{\Var} (\widehat{\omega}_j(\widehat{\cal S}_t)  X_j | X_{\widehat{\cal S}_t  }) \big ]     }{  \widehat{ E} [\widehat{\omega}_j(\widehat{\cal S}_t)  X_j  ]   },
\end{align}
where   $\widehat{ E} \big [ \widehat{\Var} (\widehat{\omega}_j(\widehat{\cal S}_t)  X_j| X_{\widehat{\cal S}_t  }) \big ]  =  \widehat{E}  \Big [    \widehat{\omega}_j^2(\widehat{\cal S}_t)  \big (  \widehat{E} [ X_j^2| X_{\widehat{\cal S}_t  } ]   -   ( \widehat{E} [X_j| X_{\widehat{\cal S}_t  } ]      )^2    \big )     \Big ]$, $\widehat{\omega}_j(\widehat{\cal S}_t) = (\beta_{j1} + \beta_{j2} \widehat{E}[ X_j | X_{\widehat{\cal S}_t}] )^{-1}$.
By Theorem \ref{thm1}, ${\cal A}_t$ can be  estimated as
$\widehat{\cal A}_{t}=\big \{j, \ |\widehat{\cal R}(j, \widehat{\cal S}_t )-1| \leq \epsilon_t \big \}$ for  some small positive constant $\epsilon_t$. This procedure is repeated until there are no remaining nodes, and then the topological layers of the DAG model are reconstructed. Note that the details for estimating $\widehat{\cal R}(j, \emptyset)$ and $\widehat{\cal R}(j, \widehat{\cal S}_t )$ may vary from one distribution to another, and we illustrate some details for both Poisson and Binomial DAGs in Appendix I, where $\widehat{E}[X_j | X_{\widehat{\cal S}_t}]$ is estimated via the  generalized linear model (GLM). The computation of $\widehat{\cal R}(j, \widehat{\cal S}_t )$ involves two terms,  $\widehat{E}[ X_j | X_{\widehat{\cal S}_t}]$ and $\widehat{E}[ X_j^2 | X_{\widehat{\cal S}_t}]$, whose computational details are provided in Appendix I. 

Once all the topological layers are reconstructed, the directed edges between nodes can be recovered via standard sparse regression models \citep{Meinshausen,yang2015} in a parallel fashion. Specifically,  for each node $j \in \widehat{\cal A}_t$, we conduct a sparse regression of $X_j$ against $X_{\widehat{\cal S}_t}$, and any non-zero coefficient leads to a directed edge pointing from the corresponding node in $\widehat{\cal S}_t$ to the node $j$. This sparse regression procedure can be done for all nodes simultaneously to expedite the computation.

The proposed two-step learning algorithm for a QVF-DAG via topological layer is summarized in  Algorithm  \ref{alg:1}, denoted as the TLDAG algorithm. 
	\begin{algorithm}[H]
	\caption{ }
	\label{alg:1}
	\begin{algorithmic}[1]
		\STATE Input: sample matrix $X^n \in {\cal R}^{n \times p}$, $\widehat{\cal S}=\emptyset$, and $t=0$;
		\STATE Until  $\widehat{\cal S}={\cal V}$:
		\begin{itemize}
			\item[a.]	 For any $j \in   {\cal V}\backslash \widehat{\cal S}$,  compute the ratio $
			\widehat{\cal R}(j, \widehat{\cal S}_t )=\frac{  \widehat{ E} \big [ \widehat{\Var} (\widehat{\omega}_j(\widehat{\cal S}_t)  X_j| X_{\widehat{\cal S}_t  }) \big ]     }{  \widehat{ E} [\widehat{\omega}_j(\widehat{\cal S}_t)  X_j  ]   }$;
			\item[b.]	Define  $
			\widehat{\cal A}_{t}=\Big \{j, \ |\widehat{\cal R}(j, \widehat{\cal S}_t )-1| \leq \epsilon_t \Big \}
			$ and let $\widehat{\cal S}=\widehat{\cal S} \cup \widehat{\cal A}_{t}$;
			\item[c.] $t\leftarrow t+1$.
		\end{itemize}
		\STATE	Let $\widehat{T}=t$.
		\STATE For any node $j \in \widehat{\cal A}_t$, fit a sparse regression model of $X_j$ against $X_{\widehat{\cal S}_t}$ to obtain the estimated directed edges pointing to node $j$, denoted as  $\widehat{\cal E}_j = \{ k \to j | k \in \widehat{\cal S}_t   \}$.\\
		\STATE Return: $\{\widehat{\cal A}_t\}_{t=0}^{\widehat{T}-1}$ and
$\{ \widehat{\cal E}_j \}_{j= \widehat{\cal A}_1}^{\widehat{\cal A}_{\widehat{T}-1}}$.
	\end{algorithmic}
\end{algorithm}

Note that Algorithm \ref{alg:1} can be modified to deal with the case with large $p$,  by replacing the GLM model in Step  2 with the $\ell_1$-regularized GLM model. It is also worth pointing out that the asymptotic consistency of the TLDAG algorithm can be established following a similar treatment as in \cite{Park20191} and \cite{SunWW2013}, with some slight modification by involving  the concept of topological layers and considering the stability selection for  $\epsilon_t$. More precisely, the assumption similar to \cite{Park20191} assumes that the true ratio for all the nodes contained in the lower layers must be bounded away from $1+\eta_{\text{min}}$ for some $\eta_{\text{min}} > 0$. Moreover, following the proof of Theorem 1 in \cite{SunWW2013}, we can show that  the reconstructed topological layers with   $\epsilon_t$  selected by the  stability selection procedure  is exactly the same as the true topological layers with high probability. Then,  sparse regression, such as $\ell_1$-regularized regression, can be applied  among each topological layer  to recover directed structures among the nodes in the layers, and the selection consistency can also be established under mild conditions as in the literature of sparse regression. Combining the above results, the consistency of the proposed algorithm can be established in the sense that it can exactly recover the true  DAG with high probability.

\subsection{Computational complexity}

In literature, the ODS algorithm  \citep{Park20191} is developed for learning QVF-DAG, and the moments ratio scoring algorithm (MRS; \cite{Park20192}) extends the ODS algorithm for a special type of QVF-DAGs, yet the computational complexities of both algorithms are of order $O(np^2)$ and $O(np^3)$, respectively, which  are still relatively high and difficult to scale up for large-scale QVF-DAGs.

By contrast, the computational complexity of the proposed TLDAG algorithm can be much less than that of ODS and MRS, especially when dealing with shallow QVF-DAGs with $T \ll p$. For example, to learn a hub graph with $T=2$ as in Figure \ref{fig:hub}, TLDAG needs to first identify ${\cal A}_0$ and ${\cal A}_1$ and then reconstruct the parent-child relationship. More precisely, identifying ${\cal A}_0$ requires estimation of the unconditional ratio for $p$ nodes, which amounts to the complexity of order $O(np)$, and identifying ${\cal A}_1$ requires estimation of the conditional ratio for the remaining $p-1$ nodes via GLM with only one predictor in ${\cal A}_0$, which also amounts to the complexity of order $O(np)$. For parent-child reconstruction, we just need to fit the GLM model for each node in ${\cal A}_1$ against the only predictor in ${\cal A}_0$, which amounts to the complexity of order $O(np)$. Therefore, the computational complexity of TLDAG is only of order $O(np)$, which is much more efficient than both ODS and MRS algorithms.  

In general, the complexity of TLDAG in estimating a random graph with $T$ layers is of order $O\Big (np(T-1)+\sum_{t=1}^{T-1}n(\sum_{k=0}^{t-1}a_k)a_t   \Big )$,  where  $a_t$ denotes  the number of nodes in   ${\cal A}_t$ and $\sum_{t=0}^{T-1}a_t=p$. Clearly, TLDAG needs to first reconstruct the topological layers $\{ {\cal A}_t \}_{t=0}^{T-1}$ in a sequential fashion, and then conduct $\ell_1$-regularized GLM regressions to recover the parent-child relation.  Specifically,  for reconstructing the topological layers, the total number of ratios needed to be computed is bounded by $O(p(T-1))$, and each ratio calculation requires $n$ samples. Thus, the complexity for reconstructing all layers is of order $O(np(T-1))$. Furthermore,  the directed parent-child structure can be recovered in a parallel fashion, and for each node,  $\ell_1$-regularized GLM regression is fitted using coordinate descent \citep{Friedman2010, Park20192}, with all the nodes in the upper layers as predictors, leading to the total complexity of order $O\big (\sum_{t=1}^{T-1}n(\sum_{k=0}^{t-1}a_k)a_t \big )$. It is clear that the computational complexity of TLDAG is the same as that of ODS when $T=p$, and it can be significantly better than ODS when the QVF-DAG has a shallow structure with $T < p$. 


\subsection{Tuning}\label{sec:tuning}

The numerical  performance of the proposed TLDAG algorithm depends on the layer reconstruction parameter $\epsilon_t$ and the tuning parameter in the sparse regression algorithm. Whereas the latter can be determined via cross-validation, we need to modify the stability-based criterion in \cite{SunWW2013} to select the optimal  parameter $\epsilon_t$ for each layer.

The  key idea is to measure the reconstruction stability by randomly splitting the training sample into two parts and comparing the disagreement between the two estimated active sets.
Specifically, given a  value $\epsilon_t$, we randomly split the training sample ${\cal Z}^M$ into two parts ${\cal Z}^M_1$ and ${\cal Z}^M_2$. Then the proposed method is applied to ${\cal Z}^M_1$ and ${\cal Z}^M_2$ and obtain two estimated active sets  $\widehat{\cal A}_{1,\epsilon_t}$ and $\widehat{\cal A}_{2,\epsilon_t}$, respectively. The disagreement between $\widehat{\cal A}_{1,\epsilon_t}$ and $\widehat{\cal A}_{2,\epsilon_t}$ is measured by Cohen's kappa coefficient
$$
\kappa(\widehat{\cal A}_{1,\epsilon_t},\widehat{\cal A}_{2,\epsilon_t})=\frac{Pr(a)-Pr(e)}{1-Pr(e)},
$$
where $Pr(a)=\frac{n_{11}+n_{22}}{p_n}$ and  $Pr(e)=\frac{(n_{11}+n_{12})(n_{11}+n_{21})}{p_n^2}+\frac{(n_{12}+n_{22})(n_{21}+n_{22})}{p_n^2}$ with $n_{11}=|\widehat{\cal A}_{1,\epsilon_t} \cap\widehat{\cal A}_{2,\epsilon_t}|, n_{12}=|\widehat{\cal A}_{1,\epsilon_t} \cap \widehat{\cal A}^C_{2,\epsilon_t}|,n_{21}=|\widehat{\cal A}^C_{1,\epsilon_t} \cap \widehat{\cal A}_{2,\epsilon_t}|, n_{22}=|\widehat{\cal A}^C_{1,\epsilon_t} \cap \widehat{\cal A}^C_{2,\epsilon_t}|$ and $|\cdot|$ denotes the set cardinality. The procedure is repeated for $B$ times and the estimated reconstruction stability is measured as
$$
\hat s(\Psi_{\epsilon_t})=\frac{1}{B} \sum_{b=1}^B \kappa(\widehat{\cal A}_{1,\epsilon_t}^b, \widehat{\cal A}_{2,\epsilon_t}^b),
$$
where $\widehat{\cal A}_{1,\epsilon_t}^b$ and $\widehat{\cal A}_{2,\epsilon_t}^b$ are the estimated active sets in the $b$-th
splitting. Finally, we set $\epsilon_t = \min \big \{ \epsilon_t: \frac{\hat s(\Psi_{\epsilon_t})}{\max_{\epsilon_t}\hat s(\Psi_{\epsilon_t})} \geq c \big \}$, where $c \in (0,1)$ is some given percentage. For illustration, $c=0.9$ and $B=5$ are used in all the numerical examples, and yield satisfactory performance.

\section{Simulated experiments}\label{sec4}

In this section,  we examine the numerical performance of the proposed TLDAG algorithm,  and  compare it against some state-of-the-art methods in terms of estimation accuracy of directed edges and computational efficiency. Specifically, five competitors are considered, including the ODS algorithm, the MRS algorithm, a direct linear non-Gaussian DAG method (DLiNGAM; \cite{shimizu2011}),  the greedy equivalence search method (GES; \cite{chickering2003}),  and the max-min hill climbing method (MMHC; \cite{Tsama}).  We implement TLDAG, ODS and MRS in R  and the source codes are available in \url{https://github.com/WeiZHOU23/TLDAG}, implementation of DLiNGAM is available on the author's website \url{https://github.com/cdt15/lingam}, and GES and MMHC are implemented in the R packages ``pcalg" \citep{Kalisch2012} and ``bnlearn" \citep{Scutari}, respectively. For GES, the output is a partial DAG, and we follow the treatment of \citealp{Yuan2019} and extend the output to DAG for fair comparison.  For TLDAG,  the tuning parameter $\epsilon_t$'s are adaptively chosen  for each layer  via the stability selection procedure in Section \ref{sec:tuning}, where the grid search is conducted over grids $\{10^{-2+0.15s}; s=0,$\ldots$,60\}$.

For comparison metrics,  we use Recall, Precision, F1-score and the normalized hamming distance (HM) to evaluate the estimation accuracy. Whereas the first three metrics are standard and popularly used in literature, HM measures the number of edge insertions, deletions or flips needed to transform one graph to another \citep{Tsama}. Large values of Recall, Precision and  F1-score and small values of HM indicate  good estimation accuracy.

In all simulated examples, we generate data for QVF-DAG models with both hub graphs and random graphs. The conditional distribution of each node given its parents follows either a Poisson distribution with rate $\exp( \theta_j + \sum_{k \in \text{pa}_j} \theta_{jk} X_k )$, or a Binomial distribution with $N_j$ trials and success rate $\text{logit}^{-1}(\theta_{j} + \sum_{k \in \text{pa}_j} \theta_{jk} X_k ))$.
A non-zero coefficient $\theta_{jk}$ indicates a directed edge from node $k$ to node $j$, and $\theta_{jk}=0$ otherwise. 

\subsection{Hub graphs}\label{sec:hub}

The hub graph is a special type of DAGs, which consists of a hub node and a number of other nodes, and directed edges only pointing from the hub node to other nodes. In this section, we consider the following hub graphs, and similar examples have also been considered in \cite{Park20192} and \cite{Yuan2019}.

\begin{figure}[!h]
	\caption{The hub graph for Examples 1 and 2.}\label{fig:hub}
	\centering
	\includegraphics[width=0.5\textwidth]{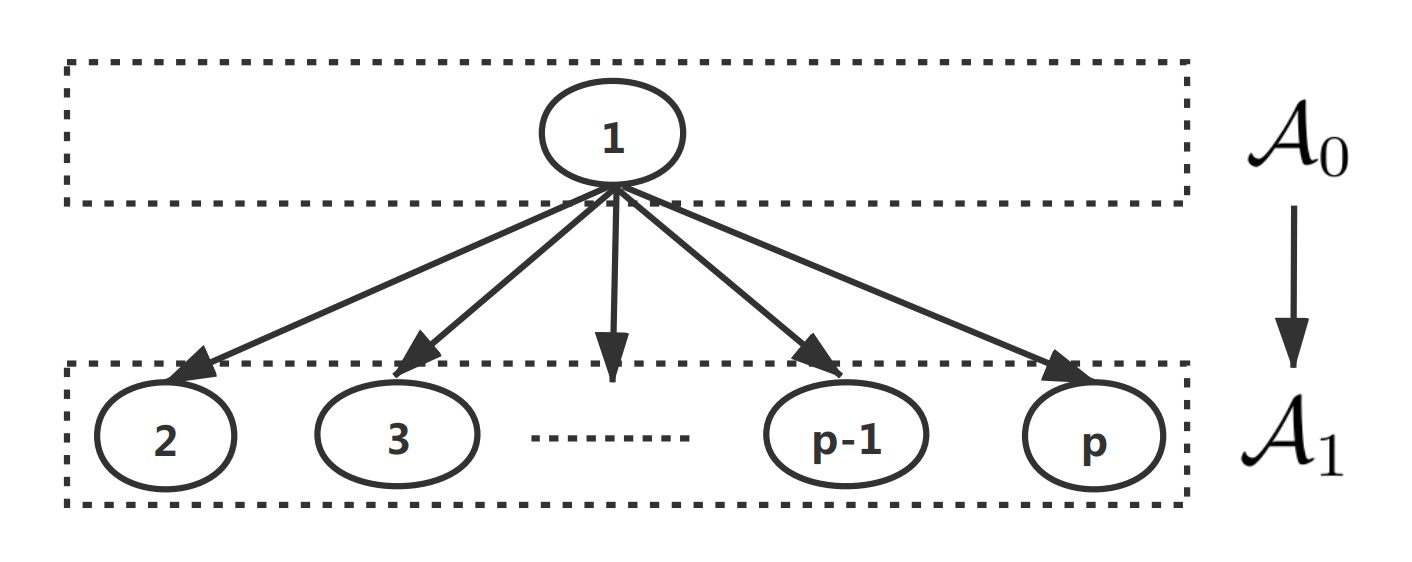}
\end{figure}

\noindent{\bf Example 1} (Poisson hub graph). The generated DAG model is depicted in Figure \ref{fig:hub}. We first generate the hub node $X_1$  in ${\cal A}_0$ from
Pois$(\exp(\theta_1))$, and then $X_j$ in ${\cal A}_1$ from Pois$(\exp( \theta_j +  \theta_{j1} X_1 ))$ for $j=2, \ldots, p$. The parameters $\theta_j$ and  $\theta_{j1}$ are generated  uniformly  from  $[1,3]$ and $[0.1, 0.5]$ for $j=1,\ldots, p$, respectively.

\noindent{\bf Example 2} (Mixed hub graph).
The generated DAG is the same as that in Example 1, but each node is generated from a mixed
distribution. Particularly, we first generate the hub node $X_1$ in ${\cal A}_0$ from
$0.5 \text{Pois}(\exp(\theta_1)) + 0.5 \text{Bin}(4 |  \theta_1 )$ and then
$X_j$ in ${\cal A}_1$ from
$0.5 \text{Pois}(\exp( \theta_j +  \theta_{j1} X_1 )) + 0.5 \text{Bin}(4 |  \theta_j +  \theta_{j1} X_1 )$ for $j=2,\ldots,p$. For $X_j$ from the Poisson distribution, the parameters $\theta_j$ and $\theta_{jk}$ are generated uniformly from $[1,3]$  and $[0.1,0.3]$, $[0.1,0.2]$ and $[0.05,0.2]$ for for $p=5$, $20$, $p=100$, respectively; and for $X_j$ from the Binomial distribution, the parameters $\theta_{j}$ and $\theta_{jk}$ are both generated uniformly from $[0.1,0.2]$ for $p=5$ and $20$, and $[0.05, 0.2]$ for $p=100$. Note that the shrunk interval for large $p$ is used to avoid large values of $\theta_j +  \theta_{j1} X_1$, which leads to a contradiction with Condition 1.

In each example, the averaged performance metrics of all the competing methods over $50$ independent replications as well as their standard errors are summarized in Tables \ref{hubpois}--\ref{hubmix}.

\begin{table}[!h]
	\normalsize
	\caption{The averaged performance metrics of various methods as well as their standard errors
		in parentheses in Example 1.}
	\centering
	\begin{tabular}{ccccccc}
		\hline\hline
		$p$ & $n$ & {Methods} & {HM} & {Recall} & Precision & F1-{score}  \\
		\hline\hline
		\multirow{12}{*}{$5$} & \multirow{6}{*}{$200$}  & $\text{TLDAG}$ & $0.13(0.01)$ & $1.00(0.00)$     & $0.63(0.01)$   & $0.76(0.01)$  \\
		&                    &$\text{ODS}$ & $0.17(0.01)$ & $1.00(0.01)$     &   $0.56(0.01)$   &  $0.71(0.01)$       \\
		&                    &$\text{MRS}$ & $0.14(0.01)$ & $0.99(0.01)$     &   $0.58(0.01)$   &  $0.73(0.01)$       \\
		&                    &$\text{DLiNGAM}$ & $0.28(0.02)$ & $0.56(0.04)$     &   $0.39(0.03)$   &  $0.45(0.03)$       \\
		&                    &$\text{GES}$ & $0.43(0.02)$  & $0.34(0.03)$  &  $0.19(0.02)$  & $0.25(0.02)$   \\
		&                    &$\text{MMHC}$ & $0.25(0.02)$  & $0.54(0.05)$  & $0.41(0.04)$   & $0.46(0.04)$  \\
		\cline{2-7}
		& \multirow{6}{*}{$500$}  & $\text{TLDAG}$ & $0.14(0.01)$ & $1.00(0.00)$   &   $0.62(0.02)$   & $0.76(0.01)$  \\
		&                       &$\text{ODS}$ & $0.17(0.01)$ & $1.00(0.00)$     &  $0.56(0.01)$   &  $0.71(0.01)$       \\
		&                    &$\text{MRS}$ & $0.13(0.01)$ & $1.00(0.01)$     &   $0.59(0.01)$   &  $0.73(0.01)$       \\
		&                      &$\text{DLiNGAM}$  & $0.25(0.02)$ & $0.59(0.04)$    & $0.43(0.03)$   &  $0.49(0.03)$       \\
		&                       &$\text{GES}$ & $0.46(0.02)$  & $0.31(0.03)$  & $0.17(0.02)$  & $0.22(0.02)$   \\
		&                       &$\text{MMHC}$ & $0.28(0.02)$  & $0.56(0.04)$  &  $0.40(0.03)$  & $0.44(0.03)$  \\
		\hline\hline
		\multirow{12}{*}{$20$} & \multirow{6}{*}{$200$}  & $\text{TLDAG}$ & $0.05(0.00)$ & $1.00(0.00)$   & $0.56(0.03)$   & $0.70(0.02)$  \\
		&                    &$\text{ODS}$ & $0.11(0.00)$ & $0.99(0.01)$     &   $0.32(0.01)$   &  $0.48(0.01)$       \\
		&                    &$\text{MRS}$ & $0.06(0.00)$ & $1.00(0.00)$     &   $0.41(0.01)$   &  $0.58(0.01)$       \\
		&                    &$\text{DLiNGAM}$ & $0.29(0.01)$ & $0.61(0.05)$     &   $0.10(0.01)$   &  $0.17(0.01)$       \\
		&                    &$\text{GES}$ & $0.18(0.00)$  & $0.32(0.02)$  &  $0.11(0.01)$  & $0.16(0.01)$   \\
		&                    &$\text{MMHC}$ & $0.09(0.00)$  & $0.12(0.01)$  & $0.12(0.01)$  & $0.12(0.01)$  \\
		\cline{2-7}
		& \multirow{6}{*}{$500$}  & $\text{TLDAG}$ & $0.05(0.00)$ & $1.00(0.00)$   &   $0.55(0.03)$   & $0.69(0.02)$  \\
		&                       &$\text{ODS}$ & $0.11(0.00)$ & $1.00(0.00)$     &  $0.32(0.01)$   &  $0.48(0.01)$       \\
		&                    &$\text{MRS}$ & $0.06(0.00)$ & $0.99(0.00)$     &   $0.43(0.01)$   &  $0.60(0.01)$       \\
		&                    &$\text{DLiNGAM}$ & $0.25(0.01)$ & $0.63(0.05)$     &   $0.12(0.01)$   &  $0.20(0.02)$       \\
		&                       &$\text{GES}$ & $0.19(0.01)$  & $0.42(0.03)$    &  $0.12(0.01)$  & $0.18(0.02)$   \\
		&                       &$\text{MMHC}$ & $0.10(0.00)$  & $0.18(0.01)$   &  $0.13(0.01)$  & $0.15(0.01)$  \\
		\hline\hline
		\multirow{12}{*}{$100$} & \multirow{6}{*}{$200$}  & $\text{TLDAG}$ & $0.02(0.00)$ & $0.99(0.00)$   &  $0.44(0.03)$   & $0.59(0.02)$  \\
		&                    &$\text{ODS}$ & $0.05(0.00)$ & $0.94(0.01)$     &  $0.16(0.00)$   &  $0.28(0.01)$       \\
		&                    &$\text{MRS}$ & $0.02(0.00)$ & $0.94(0.01)$     &   $0.30(0.01)$   &  $0.46(0.01)$       \\
		&                    &$\text{DLiNGAM}$ & $0.28(0.00)$ & $0.54(0.04)$     &  $0.02(0.00)$   &  $0.04(0.00)$       \\
		&                    &$\text{GES}$ & $0.05(0.00)$  & $0.09(0.01)$  &  $0.02(0.00)$  & $0.03(0.00)$   \\
		&                    &$\text{MMHC}$ & $0.02(0.00)$  & $0.01(0.00)$  & $0.01(0.00)$  & $0.01(0.00)$  \\
		\cline{2-7}
		& \multirow{6}{*}{$500$}  & $\text{TLDAG}$ & $0.02(0.00)$ & $1.00(0.00)$   &   $0.45(0.03)$   & $0.60(0.02)$  \\
		&                       &$\text{ODS}$ & $0.05(0.00)$ & $0.97(0.00)$     &  $0.17(0.00)$   &  $0.29(0.00)$       \\
		&                    &$\text{MRS}$ & $0.02(0.00)$ & $0.98(0.00)$     &   $0.33(0.01)$   &  $0.49(0.01)$       \\
		&                    &$\text{DLiNGAM}$ & $0.26(0.00)$ & $0.59(0.03)$     &  $0.02(0.00)$   &  $0.04(0.00)$       \\
		&                       &$\text{GES}$ & $0.07(0.00)$  & $0.17(0.01)$  & $0.03(0.00)$  & $0.05(0.00)$   \\
		&                       &$\text{MMHC}$ & $0.02(0.00)$  & $0.03(0.00)$  & $0.03(0.00)$  & $0.03(0.00)$  \\
		\hline\hline
	\end{tabular}
	\label{hubpois}
	\centering
\end{table}

\begin{table}[!h]
	\normalsize
	\caption{The averaged performance metrics of various methods as well as their standard errors
		in parentheses in Example 2.}
	\centering
	\begin{tabular}{ccccccc}
		\hline\hline
		$p$ & $n$ & {Methods} & {HM} & {Recall}  & Precision & F1-{score}  \\
		\hline \hline
		\multirow{10}{*}{$5$} & \multirow{5}{*}{$200$}  & $\text{TLDAG}$ & $0.23(0.02)$ & $0.57(0.04)$     &  $0.46(0.03)$   &  $0.50(0.03)$  \\
		&                    &$\text{ODS}$ & $0.33(0.01)$  & $0.36(0.04)$  & $0.26(0.03)$  & $0.29(0.03)$       \\
		&                    &$\text{DLiNGAM}$ & $0.37(0.02)$ & $0.47(0.05)$ &  $0.27(0.03)$   &  $0.34(0.03)$       \\
		&                    &$\text{GES}$ & $0.31(0.01)$  & $0.41(0.03)$  & $0.30(0.02)$  & $0.34(0.02)$   \\
		&                    &$\text{MMHC}$ & $0.22(0.02)$  & $0.55(0.04)$  & $0.45(0.03)$  & $0.49(0.03)$  \\
		\cline{2-7}
		& \multirow{5}{*}{$500$}  & $\text{TLDAG}$ & $0.24(0.02)$ & $0.58(0.04)$   &  $0.47(0.04)$   & $0.50(0.04)$  \\
		&                       &$\text{ODS}$ & $0.33(0.01)$ & $0.39(0.04)$     &  $0.26(0.03)$   &  $0.31(0.03)$       \\
		&                    &$\text{DLiNGAM}$ & $0.34(0.02)$ & $0.55(0.05)$ &  $0.32(0.03)$   &  $0.40(0.03)$    \\
		&                       &$\text{GES}$ & $0.31(0.01)$  & $0.45(0.03)$  & $0.31(0.02)$  & $0.37(0.03)$   \\
		&                       &$\text{MMHC}$ & $0.23(0.02)$  & $0.60(0.04)$  & $0.45(0.03)$  & $0.51(0.03)$  \\
		\hline\hline
		\multirow{10}{*}{$20$} & \multirow{5}{*}{$200$} & $\text{TLDAG}$ & $0.12(0.01)$ & $0.27(0.03)$     &   $0.18(0.02)$   &  $0.21(0.02)$       \\
		&                    &$\text{ODS}$ & $0.12(0.01)$ & $0.23(0.03)$     &   $0.10(0.01)$   &  $0.13(0.02)$       \\
		&                    &$\text{DLiNGAM}$ & $0.24(0.01)$ & $0.27(0.03)$ &  $0.06(0.01)$   &  $0.10(0.01)$    \\
		&                    &$\text{GES}$ & $0.11(0.00)$  & $0.29(0.01)$  & $0.17(0.01)$  & $0.21(0.01)$   \\
		&                    &$\text{MMHC}$ & $0.09(0.00)$  & $0.12(0.01)$  &  $0.11(0.01)$  & $0.11(0.01)$  \\
		\cline{2-7}
		& \multirow{5}{*}{$500$}  & $\text{TLDAG}$ & $0.12(0.01)$ & $0.31(0.03)$   &  $0.20(0.03)$   & $0.23(0.02)$  \\
		&                       &$\text{ODS}$ & $0.14(0.01)$ & $0.27(0.03)$     & $0.11(0.01)$   &  $0.14(0.02)$       \\
		&                    &$\text{DLiNGAM}$ & $0.25(0.01)$ & $0.33(0.03)$ &  $0.08(0.01)$   &  $0.13(0.01)$    \\
		&                       &$\text{GES}$ & $0.10(0.00)$  & $0.37(0.02)$  & $0.20(0.01)$  & $0.26(0.01)$   \\
		&                       &$\text{MMHC}$ & $0.09(0.00)$  & $0.22(0.01)$  & $0.17(0.01)$  & $0.19(0.01)$  \\
		\hline\hline
		\multirow{10}{*}{$100$} & \multirow{5}{*}{$200$}  & $\text{TLDAG}$ & $0.05(0.00)$ & $0.22(0.02)$   & $0.09(0.02)$   & $0.12(0.02)$  \\
		&                    &$\text{ODS}$ & $0.06(0.00)$ & $0.21(0.02)$     &  $0.04(0.01)$   &  $0.07(0.01)$       \\
		&                    &$\text{DLiNGAM}$ & $0.18(0.01)$ & $0.40(0.02)$ &  $0.02(0.00)$   &  $0.04(0.00)$    \\
		&                    &$\text{GES}$ & $0.03(0.00)$  & $0.12(0.01)$  &  $0.05(0.00)$  & $0.07(0.01)$   \\
		&                    &$\text{MMHC}$ & $0.02(0.00)$  & $0.02(0.00)$  & $0.01(0.00)$  & $0.02(0.00)$  \\
		\cline{2-7}
		& \multirow{5}{*}{$500$}  & $\text{TLDAG}$ & $0.05(0.00)$ & $0.29(0.03)$   &  $0.10(0.02)$   & $0.14(0.02)$  \\
		&                       &$\text{ODS}$ & $0.07(0.00)$ & $0.26(0.02)$     & $0.05(0.01)$   &  $0.07(0.01)$       \\
		&                    &$\text{DLiNGAM}$ & $0.15(0.01)$ & $0.31(0.02)$ &  $0.02(0.00)$   &  $0.04(0.00)$    \\
		&                       &$\text{GES}$ & $0.04(0.00)$  & $0.19(0.02)$  & $0.06(0.01)$  & $0.10(0.01)$   \\
		&                       &$\text{MMHC}$ & $0.02(0.00)$  & $0.03(0.00)$  & $0.02(0.00)$  & $0.03(0.00)$  \\
		\hline \hline
	\end{tabular}
	\label{hubmix}
\end{table}

It is evident that TLDAG yields superior numerical performance and outperforms the other five competitors in almost all the scenarios. In Table \ref{hubpois} with the Poisson hub graph, TLDAG yields a small HM and the largest Precision and F1-score, and the recalls of TLDAG, ODS and MRS are all close to $1$, but the other three methods have much smaller recalls.  In Table \ref{hubmix} with the mixed hub graph, TLDAG yields the best performance in terms of Precision and F1-score, and comparable performance to the best performer in terms of HM and Recall.

\subsection{Random graphs}\label{sec.ran}

We now consider two commonly used models for the random graphs, including the Erd\"os and R\'enyi (ER) model \citep{erdHos1960evolution}  and the Barab\'asi-Albert (BA)   model \citep{barabasi1999emergence}.  It is interesting to note that the BA model generates  scale-free graphs, which commonly appears in many science problems, such as the gene networks.

\begin{figure}[!h]
	\normalsize
	\caption{The random graph for  Examples 3 and 4.}\label{fig:random}
	\centering
	\includegraphics[width=0.5\textwidth]{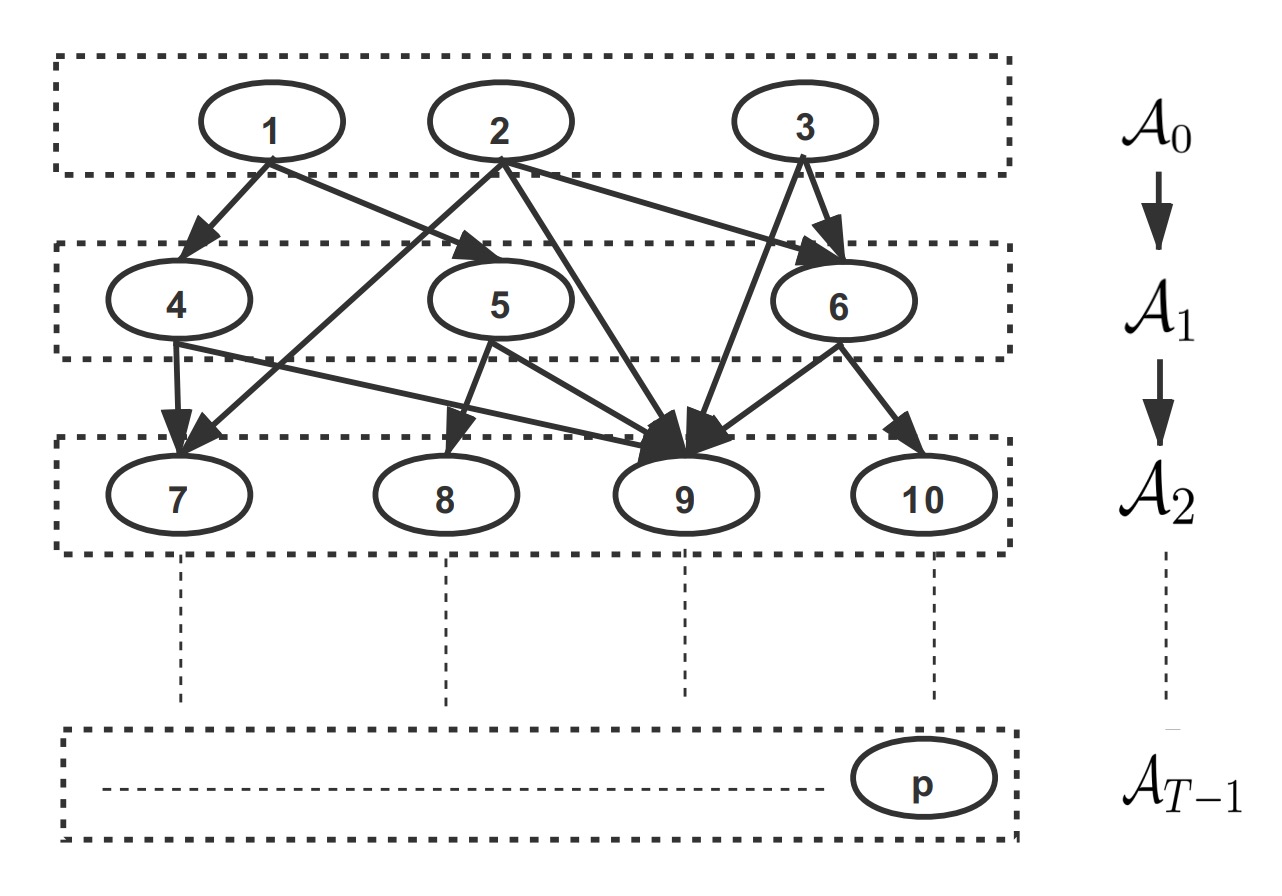}
\end{figure}

\noindent{\bf Example 3} (Mixed ER graph).  The generated DAG model is depicted in Figure  \ref{fig:random}. We set the probability of connecting an edge  as $P_E= 0.35$ for $p=5$ and $20$, and $P_E=0.1$ for $p=100$ and  generate a random DAG. Then, we convert the generated random DAG into topological structure, and  generate the data for the root nodes $X_j$'s from $0.5 \text{Pois}(\exp(\theta_j)) + 0.5 \text{Bin}(4 | \theta_j)$ and the remaining nodes $X_k$'s from $0.5 \text{Pois}(\exp(\theta_k + \sum_{l \in \text{pa}_k} \theta_{kl} X_l )) + 0.5 \text{Bin}(4 | \theta_k + \sum_{l \in \text{pa}_k} \theta_{kl} X_l )$.   Precisely,  the parameter $\theta_{k}$ for the Poisson distribution is generated uniformly from $[1, 3]$,
and $\theta_{kl}$ are generated uniformly from $[0.01, 0.05]$, $[0.005, 0.015]$, $[0.001, 0.01]$ for
$p=\{5,20,100\}$ respectively. The parameter $\theta_{k}$ and  $\theta_{kl}$ for the Binomial distribution are all generated uniformly from $[0.01, 0.05]$, $[0.005, 0.015]$, $[0.005, 0.01]$ for
$p=\{5,20,100\}$ respectively.

\noindent  {\bf Example 4} (Mixed BA graph). The generated DAG is the same as that in Example 3 except that   we set the number of edges to be added  as $e=2$ for the BA model. Additionally, the parameter $\theta_{k}$ for the Poisson distribution is generated uniformly from $[1,3]$, and $\theta_{kl}$ are generated uniformly from $[0.01, 0.03]$, $[0.005, 0.02]$, $[0.001, 0.01]$ for $p=\{5,20,100\}$ respectively. The parameter $\theta_{k}$ and  $\theta_{kl}$ for the Binomial distribution are all generated uniformly from $[0.01, 0.05]$, $[0.005, 0.02]$, $[0.001, 0.01]$ for $p=\{5,20,100\}$ respectively.

In each example, the averaged performance metrics of all the competing methods over $50$ independent replications as well as their standard errors are summarized in Tables  \ref{ran_ER}--\ref{ran_BA}.   Note that  the averaged number of topological layers considered in Example 3 and 4  for the case $p=100$ are as large as $18$ and $9$, respectively.

\begin{table}[!h]\footnotesize
	\caption{The averaged performance metrics of various methods as well as their standard errors
		in parentheses in Example 3.}
	\normalsize
	\centering
	\begin{tabular}{ccccccc}
		\hline \hline
		$p$ & $n$ & $\text{Methods}$ & $\text{HM}$ & $\text{Recall}$  & Precision & F1-{score}  \\
		\hline\hline
		\multirow{10}{*}{$5$} & \multirow{5}{*}{$200$}  & $\text{TLDAG}$ & $0.27(0.02)$ & $0.34(0.04)$     &   $0.34(0.05)$   &  $0.31(0.04)$  \\
		&                    &$\text{ODS}$ & $0.32(0.01)$  & $0.25(0.03)$  & $0.22(0.04)$  & $0.22(0.03)$       \\
		&                    &$\text{DLiNGAM}$ & $0.41(0.02)$ & $0.39(0.05)$ &  $0.19(0.02)$   &  $0.24(0.03)$    \\
		&                    &$\text{GES}$ & $0.31(0.01)$  & $0.37(0.04)$  &  $0.25(0.03)$  & $0.28(0.03)$   \\
		&                    &$\text{MMHC}$ & $0.29(0.01)$ & $0.28(0.03)$  & $0.26(0.03)$  & $0.25(0.03)$  \\
		\cline{2-7}
		& \multirow{5}{*}{$500$}  & $\text{TLDAG}$ & $0.24(0.02)$ & $0.39(0.04)$   &   $0.41(0.05)$   & $0.37(0.04)$  \\
		&                       &$\text{ODS}$ & $0.32(0.01)$ & $0.29(0.03)$     & $0.24(0.03)$   &  $0.25(0.03)$       \\
		&                    &$\text{DLiNGAM}$ & $0.39(0.02)$ & $0.40(0.05)$ &  $0.20(0.02)$   &  $0.25(0.03)$    \\
		&                       &$\text{GES}$ & $0.31(0.01)$  & $0.38(0.05)$  & $0.25(0.03)$  & $0.28(0.03)$   \\
		&                       &$\text{MMHC}$ & $0.29(0.01)$  & $0.33(0.04)$  & $0.26(0.03)$  & $0.27(0.03)$  \\
		\hline\hline
		\multirow{10}{*}{$20$} & \multirow{5}{*}{$200$}  & $\text{TLDAG}$ & $0.20(0.01)$ & $0.12(0.01)$   & $0.37(0.03)$   & $0.16(0.01)$  \\
		&                    &$\text{ODS}$ & $0.23(0.00)$ & $0.12(0.01)$     &  $0.22(0.01)$   &  $0.15(0.01)$       \\
		&                    &$\text{DLiNGAM}$ & $0.31(0.01)$ & $0.27(0.02)$ &  $0.20(0.01)$   &  $0.22(0.01)$    \\
		&                    &$\text{GES}$ & $0.23(0.00)$  & $0.17(0.01)$  & $0.26(0.01)$  & $0.20(0.01)$   \\
		&                    &$\text{MMHC}$ & $0.21(0.00)$  & $0.07(0.01)$  & $0.20(0.01)$  & $0.11(0.01)$  \\
		\cline{2-7}
		& \multirow{5}{*}{$500$}  & $\text{TLDAG}$ & $0.21(0.01)$ & $0.14(0.01)$   & $0.36(0.03)$   & $0.17(0.02)$  \\
		&                       &$\text{ODS}$ & $0.23(0.01)$ & $0.13(0.01)$     &$0.22(0.02)$   &  $0.15(0.01)$       \\
		&                    &$\text{DLiNGAM}$ & $0.31(0.01)$ & $0.27(0.02)$ &  $0.21(0.01)$   &  $0.22(0.01)$    \\
		&                       &$\text{GES}$ & $0.23(0.00)$  & $0.22(0.02)$  &$0.27(0.01)$  & $0.24(0.01)$   \\
		&                       &$\text{MMHC}$ & $0.21(0.00)$  & $0.11(0.01)$  & $0.24(0.01)$  & $0.15(0.01)$  \\
		\hline \hline
		\multirow{10}{*}{$100$} & \multirow{5}{*}{$200$}  & $\text{TLDAG}$ & $0.07(0.00)$ & $0.03(0.00)$   &  $0.15(0.02)$   & $0.04(0.00)$  \\
		&                    &$\text{ODS}$ & $0.10(0.00)$ & $0.05(0.01)$     &  $0.05(0.00)$   &  $0.05(0.00)$       \\
		&                    &$\text{DLiNGAM}$ & $0.24(0.01)$ & $0.24(0.01)$ &  $0.06(0.00)$   &  $0.08(0.00)$    \\
		&                    &$\text{GES}$ & $0.08(0.00)$  & $0.09(0.00)$  & $0.11(0.00)$  & $0.10(0.00)$   \\
		&                       &$\text{MMHC}$ & $0.06(0.00)$  & $0.02(0.00)$  & $0.08(0.00)$  & $0.03(0.00)$ \\
		\cline{2-7}
		& \multirow{5}{*}{$500$}  & $\text{TLDAG}$ & $0.08(0.00)$ & $0.06(0.01)$   &  $0.10(0.01)$   & $0.06(0.00)$  \\
		&                       &$\text{ODS}$ & $0.12(0.00)$ & $0.08(0.01)$     &   $0.05(0.00)$   &  $0.06(0.00)$       \\
		&                    &$\text{DLiNGAM}$ & $0.18(0.00)$ & $0.18(0.01)$ &  $0.06(0.00)$   &  $0.09(0.00)$    \\
		&                       &$\text{GES}$ & $0.09(0.00)$  & $0.14(0.00)$  &  $0.13(0.00)$  & $0.14(0.00)$   \\
		&                       &$\text{MMHC}$ & $0.07(0.00)$  & $0.04(0.00)$  & $0.10(0.00)$  & $0.05(0.00)$ \\
		\hline
		\hline
	\end{tabular}
	\label{ran_ER}
\end{table}

\begin{table}[!h]
	\normalsize
	\caption{The averaged performance metrics of various methods as well as their standard errors
		in parentheses in Example 4.}
	\centering
	\begin{tabular}{ccccccc}
		\hline\hline
		$p$ & $n$ & $\text{Methods}$ & $\text{HM}$ & $\text{Recall}$  & Precision & F1-{score}  \\
		\hline\hline
		\multirow{10}{*}{$5$} & \multirow{5}{*}{$200$} &$\text{TLDAG}$ & $0.28(0.02)$  & $0.40(0.03)$  & $0.73(0.05)$  & $0.49(0.04)$       \\
		&                    &$\text{ODS}$ & $0.35(0.02)$  & $0.36(0.03)$  & $0.49(0.04)$  & $0.41(0.03)$   \\
		&                    &$\text{DLiNGAM}$ & $0.47(0.02)$ & $0.38(0.03)$ &  $0.35(0.03)$   &  $0.36(0.03)$    \\
		&                    &$\text{GES}$ & $0.39(0.01)$ & $0.33(0.03)$  & $0.44(0.05)$  & $0.37(0.04)$  \\
		&                    &$\text{MMHC}$ & $0.42(0.01)$ & $0.23(0.02)$  & $0.35(0.05)$  & $0.27(0.03)$  \\
		\cline{2-7}
		& \multirow{5}{*}{$500$}  & $\text{TLDAG}$ & $0.28(0.01)$ & $0.47(0.03)$   & $0.70(0.04)$   & $0.54(0.03)$  \\
		&                       &$\text{ODS}$ & $0.37(0.02)$ & $0.35(0.03)$     &  $0.47(0.03)$   &  $0.39(0.03)$       \\
		&                    &$\text{DLiNGAM}$ & $0.45(0.02)$ & $0.39(0.04)$ &  $0.37(0.04)$   &  $0.37(0.03)$    \\
		&                       &$\text{GES}$ & $0.40(0.02)$  & $0.36(0.04)$  & $0.42(0.05)$  & $0.38(0.04)$   \\
		&                       &$\text{MMHC}$ & $0.43(0.01)$  & $0.25(0.03)$  & $0.34(0.04)$  & $0.29(0.03)$  \\
		\hline\hline
		\multirow{10}{*}{$20$} & \multirow{4}{*}{$200$}  & $\text{TLDAG}$ & $0.12(0.01)$ & $0.15(0.01)$   &  $0.49(0.05)$   & $0.21(0.02)$  \\
		&                    &$\text{ODS}$ & $0.18(0.00)$ & $0.08(0.02)$     &$0.09(0.02)$   &  $0.08(0.02)$       \\
		&                    &$\text{DLiNGAM}$ & $0.25(0.01)$ & $0.27(0.02)$ &  $0.12(0.01)$   &  $0.17(0.01)$    \\
		&                    &$\text{GES}$ & $0.17(0.00)$  & $0.17(0.01)$  & $0.16(0.02)$  & $0.17(0.01)$   \\
		&                    &$\text{MMHC}$ & $0.15(0.00)$  & $0.08(0.01)$  & $0.12(0.01)$  & $0.10(0.01)$  \\
		\cline{2-7}
		& \multirow{5}{*}{$500$}  & $\text{TLDAG}$ & $0.13(0.01)$ & $0.18(0.03)$   &  $0.48(0.05)$   & $0.24(0.02)$  \\
		&                       &$\text{ODS}$ & $0.19(0.01)$ & $0.08(0.02)$     &  $0.08(0.02)$   &  $0.08(0.02)$       \\
		&                    &$\text{DLiNGAM}$ & $0.25(0.01)$ & $0.28(0.03)$ &  $0.13(0.01)$   &  $0.18(0.01)$    \\
		&                       &$\text{GES}$ & $0.17(0.00)$  & $0.21(0.02)$  & $0.18(0.02)$  & $0.20(0.02)$   \\
		&                       &$\text{MMHC}$ & $0.15(0.00)$  & $0.10(0.01)$  &  $0.14(0.02)$  & $0.12(0.01)$  \\
		\hline \hline
		\multirow{10}{*}{$100$} & \multirow{5}{*}{$200$}  & $\text{TLDAG}$ & $0.03(0.00)$ & $0.04(0.01)$   &  $0.33(0.05)$    & $0.06(0.01)$  \\
		&                    &$\text{ODS}$ & $0.10(0.01)$ & $0.05(0.01)$     &  $0.01(0.01)$   &  $0.02(0.01)$       \\
		&                    &$\text{DLiNGAM}$ & $0.13(0.00)$ & $0.19(0.01)$ &  $0.03(0.00)$   &  $0.05(0.00)$    \\
		&                    &$\text{GES}$ & $0.05(0.00)$  & $0.08(0.00)$  & $0.05(0.00)$  & $0.06(0.00)$   \\
		&                    &$\text{MMHC}$ & $-$  & $-$  & $-$  & $-$ \\
		\cline{2-7}
		& \multirow{5}{*}{$500$} &$\text{TLDAG}$ & $0.04(0.01)$  & $0.06(0.01)$  &$0.29(0.04)$  & $0.08(0.01)$    \\
		&                      & $\text{ODS}$ & $0.13(0.01)$  & $0.07(0.01)$   &$0.01(0.00)$  & $0.02(0.00)$  \\
		&                    &$\text{DLiNGAM}$ & $0.12(0.00)$ & $0.17(0.02)$ &  $0.03(0.01)$   &  $0.05(0.01)$    \\
		&                       &$\text{GES}$ & $0.05(0.00)$ & $0.11(0.00)$     & $0.06(0.01)$   &  $0.08(0.00)$       \\
		&                       &$\text{MMHC}$ & $-$  & $-$  & $-$  & $-$ \\
		\hline \hline
	\end{tabular}
	\label{ran_BA}
	\centering
\end{table}
From Tables \ref{ran_ER} and \ref{ran_BA}, it is clear that TLDAG still performs well on the random graphs. Precisely, in the both two examples,  TLDAG  is the best performer in terms of HM and Precision, and  yields comparable performance to the other competitors in term of F1-score in almost all the scenarios. Note that   TLDAG achieves the highest Precision in all scenarios, largely due to the fact that  TLDAG can obtain high layer-recovery accuracy, leading to the better DAG estimation accuracy. It is worthy pointing out that  DLiNGAM achieves  the highest Recall in some cases,  due to the fact that it tends to produce a very dense graph with many false edges, leading to small Precision. Note that the MMHC method  does not return any result after running for more than 24 hours for the cases with $p = 100$  in Example 4, and thus is omitted in Table \ref{ran_BA}  correspondingly.

\subsection{Computational comparison}
	
We now turn to examine the computational efficiency of the proposed TLDAG algorithm. The averaged computing time (in seconds) of TLDAG, ODS and MRS in the Poisson hub graph and the mixed random graph with $p \in\{50, 100, 200\}$ and $n \in \{400, 500\}$ is summarized in Table \ref{tabletime}.

\begin{table}[!h]
	\centering
	\normalsize
	\caption{Comparison of TLDAG with ODS and MRS  in
		terms of averaged run-time (in seconds) in Examples 1 and 4.}
	\begin{tabular}{clcccccc}
		\hline
		\multicolumn{1}{c}{\multirow{2}[2]{*}{Graphs}}& \multicolumn{1}{c}{\multirow{2}[2]{*}{$p$}} & \multicolumn{3}{c}{$n=400$} & \multicolumn{3}{c}{$n=500$} \\
		\cline{3-5} \cline{6-8}
		& \multicolumn{1}{c}{} & TLDAG & ODS & MRS & TLDAG & ODS & MRS \\
		\hline
		\multirow{3}{*}{Example 1}& $50$ & $5.27$ & $58.23$  & $402.26$  & $9.41$  & $81.76$  & $451.11$  \\
		& $100$ & $15.75$ & $266.41$  & $4978.77$  & $30.21$ & $506.23$  & $8552.13$   \\
		& $200$  & $53.54$ & $940.85$  & $>10,000$  & $56.49$  & $1294.99$  & $>10,000$   \\
		\hline
		\multirow{3}{*}{Example 1} & $50$ & $9.83$ & $58.43$   &  $-$ & $13.40$  & $71.55$  & $-$  \\
		& $100$ & $42.88$ & $270.31$ & $-$  & $57.75$ & $316.95$  &  $-$  \\
		& $200$  & $195.52$ & $1196.32$ & $-$ &  $262.19$ & $1496.86$  &  $-$  \\
		\hline
	\end{tabular}%
	\label{tabletime}
\end{table}%
It is evident that TLDAG is much more efficient than other methods in terms of computational cost, where all the tuning procedures are taken into consideration. The computational efficiencies reported in Table \ref{tabletime} also support the computational complexity analysis in Section 3.2.

\section{Real applications}\label{sec5}

We now apply the proposed TLDAG algorithm to analyze two real examples, including an NBA
player statistics data and a cosmetic sales data.
The NBA player statistics data is publicly available in the R package ''SportsAnalytics", and
the cosmetic sales data is collected by Alibaba, one of the largest online stores in China.

\subsection{NBA player data}

The NBA player statistics data consists of a number of  statistics for 441 NBA players in the season 2009/2010.  For illustration, we focus on 18 informative statistics, including TotalMinutesPlayed, FieldGoalsMade, FieldGoalsAttempted, ThreesMade, ThreesAttempted, FreeThrowsMade, FreeThrowsAttempted, OffensiveRebounds, TotalRebounds, Assists, Steals, Turnovers, Blocks, PersonalFouls, Disqualifications, TotalPoints, Technicals and GamesStarted.

\begin{figure}[!h]
	\caption{ The estimated DAG among 18 statistics in the NBA player data.}\label{fig:basket}
	\centering
	\includegraphics[width=0.95\textwidth]{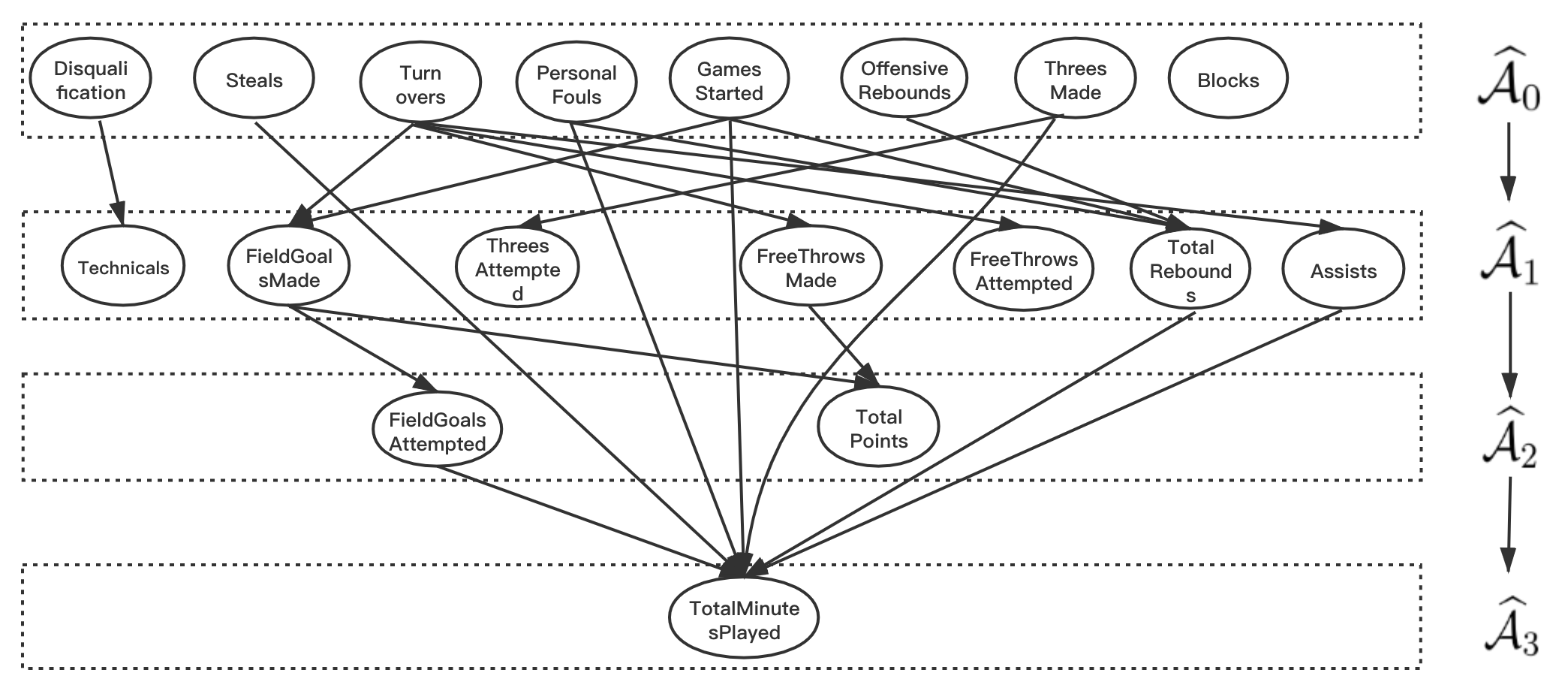}
\end{figure}

Following the same treatment as in \cite{Park20191},  we assume the conditional distribution of each node given its parents follows a Poisson distribution.  We then apply TLDAG to estimate the directed structures among the 18 statistics, as shown in Figure \ref{fig:basket}.

The estimated DAG in  Figure \ref{fig:basket} has four topological layers and  twenty directed edges.
Compared with the estimated DAG by ODS, it contains six more directed edges and reverses six directed edges, and the difference is summarized in Figure \ref{fig:basdata}.
\begin{figure}[!h]
	\normalsize
	\caption{The difference between the estimated DAGs by TLDAG and ODS.} \label{fig:basdata}
	\centering
		\begin{tabular}{c|c}
			\hline
            Added Edges  &   GamesStarted $\to$ FieldGoalsMade / TotalRebounds, \\
                         & Steals / Assists / ThreesMade $\to$  TotalMinutesPlayed, \\
                         &   Turnovers $\to$ FieldGoalsMade. \\
\hline
			Reversed Edges &  FieldGoalsMade $\to$ TotalPoints,  \\
                           &  PersonalFouls/OffensiveRebounds $\to$ TotalRebounds, \\
                           &  FieldGoalsAttempted/PersonalFouls/TotalRebounds $\to$ TotalMinutesPlayed. \\
\hline
		\end{tabular}
	\centering
\end{figure}

It is evident  that  TLDAG produces a much more reasonable DAG compared with ODS.
The added edges appear reasonable and agree with  common sense.
The more GamesStarted and the less Turnovers a player has, the more FieldGoalsMade and TotalRebounds he may obtain by controlling and dribbling more balls;
if a player has competitive ability in Steals and Assists and makes more three point shots, he is more likely to be a key player in the team and thus plays more minutes in the game.
For the reversed edges, it is reasonable that a larger number of FieldGoalMade implies more total points, but the reverse is not necessarily true;
if a player is in center or Power Forward position, he is likely to have more PersonalFouls, also more OffensiveRebounds, and thus more TotalRebouds.
Note that the position variable is seen as a latent variable and excluded in this graph.
Furthermore, strong ability of taking more TotalRebounds for a player leads to his more minutes to play,
and more FieldGoalsAttempted and less PersonalFouls show the player's offensive and defensive abilities,
which results in more minutes he plays in the game.

\subsection{Alibaba cosmetic sales data}

The cosmetic sales data set consists of  $35102$ samples and  $6$ features of liquid essences of some anonymous brands, including  the number of orders in one month (NO), the level of  brand (LB), the star level of the seller (SLS),  the place of origin (PO), the effect (EF) and the ingredient (IND).  Note that  NO is a continuous variable and the other five variables are discrete. Specifically, LB and SLS take values in $\{0,1,2,3,4,5\}$ and $\{0,1,2,3\}$, where a larger value
indicates  higher reputation for the brand and seller.
PO takes value in $\{0,1,2,\ldots,21\}$ representing $22$ different countries of origin,
EF takes value in $\{0,1,2,\ldots,31\}$ for effects including  moisurization, skin whitening, despeckle and so on, and IND takes value in $\{0,1,2,\ldots,122\}$ for different ingredients such as water, polyois, essence, solubilizer and so on.

\begin{figure}[!h]
	\caption{The estimated DAGs in the cosmetic sales data by TLDAG (left) and ODS (right).}\label{real1}
	\centering
	\includegraphics[width=0.5\textwidth]{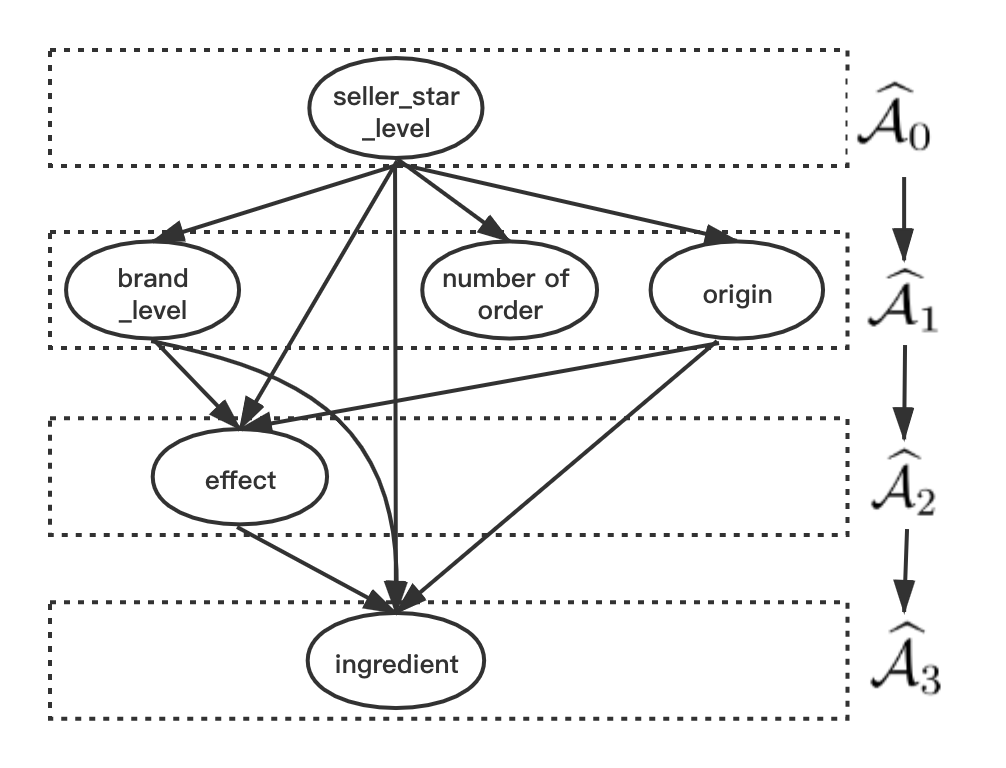}
    \includegraphics[width=0.43\textwidth]{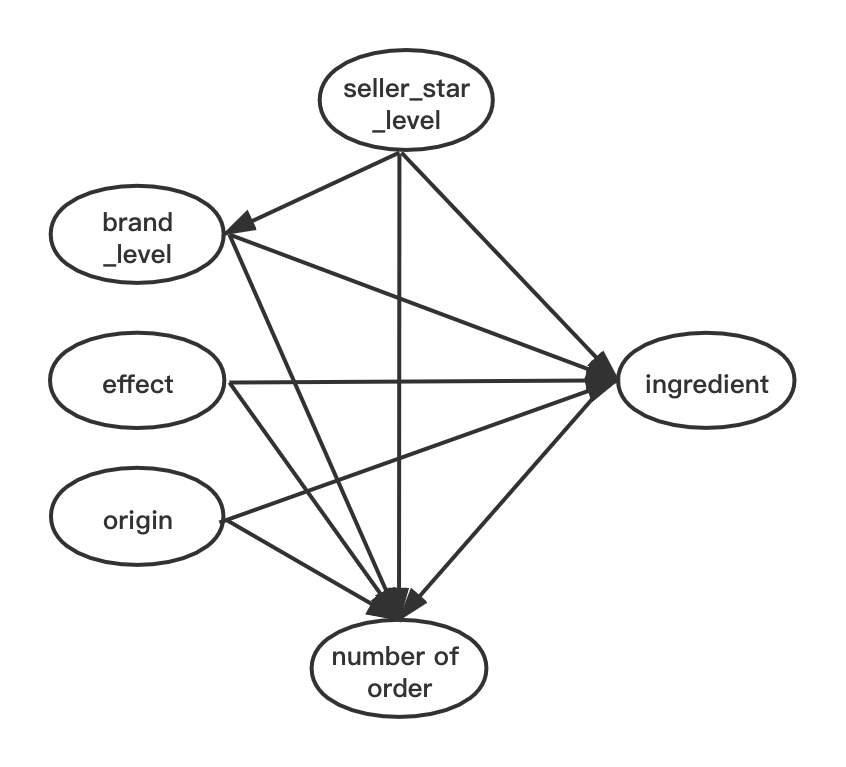}
\end{figure}

As suggested by the descriptive statistics, it is reasonable to assume that the conditional distribution
of NO given its parents follows an exponential distribution,
and the conditional distributions of the other five discrete variables
given their parents are Binomial.
We then apply the proposed  TLDAG algorithm to the dataset and the estimated DAG is shown in the left panel of  Figure \ref{real1},
which has four topological layers and ten directed edges.

In Figure \ref{real1} , some of estimated directed edges are highly interpretable.
For example, a higher star level of the seller indicates a better reputation and higher customer loyalty, which leads to a larger number of orders;
different origins have their unique materials and ingredients for the specific effect;
the higher the brand level is, the more various and advanced effects the liquid essence has.
The last two directed edges, however, are missed in the estimated DAG by ODS, shown in the right panel of Figure \ref{real1}.
It is also interesting to point out that the imposed conditional distribution assumptions  may not be satisfied by the cosmetic sales data, and thus some of estimated edges are not easy to interpret, such as the seller level $\to$ the brand level/origin.

\section{Summary}\label{sec:sum}

In this paper, we propose a computationally efficient learning algorithm for a large class of non-Gaussian DAGs, denoted as QVF-DAGs.  The proposed algorithm is based on a novel concept of topological layers, and consists of two steps of learnings. It first reconstructs the  topological layers in a hierarchical fashion, and then reconstructs the directed edges between nodes in different layers. The computational complexity of the proposed algorithm is much less than the existing learning algorithms in literature. The computational efficiency and the estimation accuracy of the proposed method are also supported by a number of simulated examples and two real applications.

\section*{Acknowledgments}
        
The authors thank the editor, the associate editor and the two anonymous referees for their constructive suggestions, which significantly improve this paper. The first two authors contribute equally to this paper. Xin He's research is supported in part by  NSFC-11901375 and Shanghai Pujiang Program
2019PJC051, Wei Zhong's research is supported in part by NSFC-11671334, NSFC-11922117
and Fujian Provincial Natural Science Fund for Distinguish Young Scholars (2019J06004)
, and Junhui Wang's research is supported in part by HK RGC Grants GRF-11303918, GRF-11300919 and GRF-11304520.

\section*{Appendix I: Computational details}

In this part, we provide some computational details for the Poisson and Binomial DAGs. For the Poisson distribution with $\beta_{j1}=1$ and $\beta_{j2}=0$, we have
\begin{eqnarray*}
\widehat{ E} [ \widehat{\omega}_j  (\widehat{\cal S}_t)  X_j  ]
 &=&
 \widehat{ E } (X_j) = \frac{1}{n} \sum_{i=1}^n  X_{i,j}^{n},
\\
 \widehat{E}  \Big [    \widehat{\omega}_j^2(\widehat{\cal S}_t)    \widehat{E} [X_j^2| X_{\widehat{\cal S}_t  } ]   \Big ]
&=&
 \frac{1}{n} \sum_{i=1}^{n}
\exp \left(   \widehat{\theta}_j^{\widehat{\cal S}_t}  +   \sum_{k \in \widehat{\cal S}_t}
\widehat{\theta}_{jk}^{\widehat{\cal S}_t}   X_{i,t}^{n}   \right),
\\
\text{and} \quad
   \widehat{E}  \Big [  \widehat{\omega}_j^2 (\widehat{\cal S}_t)   ( \widehat{E} [X_j|
   X_{\widehat{\cal S}_t  } ]      )^2     \Big ]
&=&
  \frac{1}{n}  \sum_{i=1}^{n}
\exp \left(  2  (   \widehat{\theta}_j^{\widehat{\cal S}_t} +   \sum_{k \in \widehat{\cal S}_t}
\widehat{\theta}_{jk}^{\widehat{\cal S}_t} X_{i,t}^{n}   )  \right).
\end{eqnarray*}
where $\widehat{\theta}^{\widehat{\cal S}_t}(j) = (\widehat{\theta}_j^{\widehat{\cal S}_t}, \widehat{\theta}_{j\cdot}^{\widehat{\cal S}_t} )$ is the solution of the following optimization task that
\begin{eqnarray*}
\widehat{\theta}^{\widehat{\cal S}_t}(j)
:= \argmin  \frac{1}{n}\sum_{i=1}^{n} \left(   - X_{i,j}^n  \left(  \theta_j + \sum_{k \in \widehat{\cal S}_t }
\theta_{jk} X_{i,k}^n  \right) + \exp \left(  \theta_j + \sum_{k \in \widehat{\cal S}_t }
\theta_{jk} X_{i,k}^n  \right)
\right)
\end{eqnarray*}

For the Binomial distribution with  $\beta_{j1}=1$ and $\beta_{j2}=-\frac{1}{N}$, we have
\begin{align*}
	\widehat{ E} [ \widehat{\omega}_j  (\widehat{\cal S}_t)  X_j  ]
	&=
	\frac{1}{n} \sum_{i=1}^n  \widehat{\omega}_j(\widehat{\cal S}_t)  X_{i,j}^{n},
	\\
	\widehat{E}  \Big [    \widehat{\omega}_j^2(\widehat{\cal S}_t)    \widehat{E} [X_j^2| X_{\widehat{\cal S}_t  } ]   \Big ]
	&=
	\frac{1}{n} \sum_{i=1}^{n}  \Big[  \widehat{\omega}_j(\widehat{\cal S}_t)  \widehat{p}(X_j | X_{\widehat{\cal S}_t})  (X_{i,j}^{n})^2     \Big],
	\\
	\text{and}  \quad
	\widehat{E}  \Big [  \widehat{\omega}_j^2 (\widehat{\cal S}_t)   ( \widehat{E} [ X_j|
	X_{\widehat{\cal S}_t  } ]      )^2     \Big ]
	&=
	\widehat{\omega}_j^2 (\widehat{\cal S}_t)   \left(  \frac{1}{n}  \sum_{i=1}^{n}
	\widehat{p}(X_j | X_{\widehat{\cal S}_t})  X_{i,j}^{n}            \right)^2,
\end{align*}
where  $
	\widehat{\omega}_j(\widehat{\cal S}_t)=
	\left( 1-\frac{1}{Nn} \sum_{i=1}^n \widehat{p}(X_j | X_{\widehat{\cal S}_t}) X_{i,j}^{n}  \right)^{-1}
$ with $\widehat{p}( X_j | X_{\widehat{\cal S}_t})
	=
	\frac{  \exp  \big(  \widehat{\theta}_j^{\widehat{\cal S}_t} X_{i,j}^{n} + \sum_{k \in \widehat{\cal S}_t}   \widehat{\theta}_{jk}^{\widehat{\cal S}_t}   X_{i,k}^{n} X_{i,j}^{n} \big)
		\tbinom{N_j}{X_{i,j}^{n}}  }{ \exp \big( N_j \log ( 1+\exp( \widehat{\theta}_j^{\widehat{\cal S}_t} + \sum_{k \in \widehat{\cal S}_t}
		\widehat{\theta}_{jk}^{\widehat{\cal S}_t} X_{i,k}^{n} ) ) \big)  }
$ and  $(\widehat{\theta}_j^{\widehat{\cal S}_t}, \widehat{\theta}_{j\cdot}^{\widehat{\cal S}_t} )$ denotes the solution of the following optimization task that
 \begin{eqnarray*}
\widehat{\theta}^{\widehat{\cal S}_t}(j)
&:=&
\argmin  \frac{1}{n}\sum_{i=1}^{n} \left(   - X_{i,j}^n  \left(  \theta_j + \sum_{k \in \widehat{\cal S}_t}
\theta_{jk} X_{i,k}^n  \right) + N_j \log \left( 1+ \exp \left(  \theta_j + \sum_{k \in \widehat{\cal S}_t}
\theta_{jk} X_{i,k}^n  \right) \right)
\right)
\end{eqnarray*}

Finally, the estimated ratio for each node $j=1,...,p$, can be written as
\begin{eqnarray*}
\widehat{\cal R}(j, \widehat{\cal S}_t)=
\frac{ \widehat{E}  \Big [    \widehat{\omega}_j^2(\widehat{\cal S}_t)    \widehat{E} [X_j^2| X_{\widehat{\cal S}_t  } ]   \Big ] -
\widehat{E}  \Big [  \widehat{\omega}_j^2 (\widehat{\cal S}_t)   ( \widehat{E} [X_j|X_{\widehat{\cal S}_t  } ]      )^2     \Big ]  }{\widehat{ E} [ \widehat{\omega}_j  (\widehat{\cal S}_t)  X_j  ]}.
\end{eqnarray*}

\section*{Appendix II : Proof of Lemma \ref{lemma1}}

Simple algebra yields that
\begin{align}\label{lem1:1}
	{ E \big [\Var(\omega_j({\cal S})X_j|X_{\cal S}) \big ] - E[\omega_j({\cal S})X_j  ]}&=E\Big [ \Var(\omega_j({\cal S})X_j |X_{\cal S}) -  E[ { \omega_j({\cal S}) } X_j|X_{\cal S}] \Big ] \nonumber\\
&=E\Big [  \omega_j^2({\cal S}) \big (  \Var(X_j|X_{\cal S})   - \omega^{-1}_j({\cal S})  E[X_j|X_{\cal S}] \big ) \Big ].
\end{align}
Then by total variance decomposition, \eqref{lem1:1} can be decomposed as
\begin{align*}
&E\Big [  \omega_j^2({\cal S}) \big (  \Var(X_j|X_{\cal S})   - \omega^{-1}_j({\cal S})  E[X_j|X_{\cal S}] \big ) \Big ] \\
&=E\Big [ \omega_j^2({\cal S})  \Big (  \Var( E[X_j|X_{\text{pa}_j}]|X_{\cal S}) + E[ \Var(X_j|X_{\text{pa}_j})|X_{\cal S}]   -(\beta_{j1}+ \beta_{j2} E[X_j| X_{\cal S}]   )  E[X_j|X_{\cal S}] \Big ) \Big ]\\
&=E\Big [ \omega_j^2({\cal S}) \Big (  \Var( E[X_j|X_{\text{pa}_j}]|X_{\cal S}) + \beta_{j1} E[X_j|X_{\cal S} ] + \beta_{j2} E\big [ E[ X_j|X_{\text{pa}_j}]^2|X_{\cal S} \big ]  \\
&\ \hspace{8cm}  -(\beta_{j1}+ \beta_{j2} E[X_j| X_{\cal S}]   )  E[X_j|X_{\cal S}] \Big ) \Big ]\\
& = E\Big [  \omega_j^2({\cal S}) \Big (  \Var \big ( E[X_j|X_{\text{pa}_j}]|X_{\cal S} \big ) + \beta_{j2} E\big [ E[ X_j|X_{\text{pa}_j}]^2|X_{\cal S} \big ]   -\beta_{j2} E\big [E [X_j| X_{\text{pa}_j}] | X_{\cal S} \big]^2     \Big ) \Big ]\\
& =E\Big [  \omega_j^2({\cal S})  \Big (  \Var \big ( E[X_j|X_{\text{pa}_j}]|X_{\cal S} \big ) + \beta_{j2}  \Var \big ( E[X_j|X_{\text{pa}_j}]|X_{\cal S} \big )   \Big ) \Big ]\\
&=  (1+\beta_{j2}  ) E \Big [  \omega_j^2({\cal S}) \Var \big ( E[X_j|X_{\text{pa}_j}]|X_{\cal S} \big )\Big ].
\end{align*}
where the first equality follows from \eqref{eqn:variance} and the last inequality is greater than $0$ by Condition 1 if $\text{pa}_j \nsubseteq {\cal S} \subset \text{nd}_j$, and equals $0$ when $\text{pa}_j \subseteq {\cal S} \subseteq \text{nd}_j$ and with the fact that $\beta_{j2}>-1$. This completes the proof.
\hfill $\blacksquare$

\bibliography{bibtex}

\begin{thebibliography}{32}
\providecommand{\natexlab}[1]{#1}
\providecommand{\url}[1]{\texttt{#1}}
\expandafter\ifx\csname urlstyle\endcsname\relax
  \providecommand{\doi}[1]{doi: #1}\else
  \providecommand{\doi}{doi: \begingroup \urlstyle{rm}\Url}\fi

\bibitem[Barab\'asi and Albert(1999)]{barabasi1999emergence}
A.~Barab\'asi and R.~Albert.
\newblock Emergence of scaling in random networks.
\newblock \emph{Science}, \textbf{286}:\penalty0 509--512, 1999.

\bibitem[Brown et~al.(2010)Brown, Cai, and Zhou]{Brown}
L.~D. Brown, T.~T. Cai, and H.~H. Zhou.
\newblock Nonparametric regression in exponential families.
\newblock \emph{The Annals of Statistics}, \textbf{38}:\penalty0 2005--2046,
  2010.

\bibitem[B\"uhlmann et~al.(2014)B\"uhlmann, Peters, and Ernest]{aosbuhl}
P.~B\"uhlmann, J.~Peters, and J.~Ernest.
\newblock {CAM}: {C}ausal additive models, high-dimensional order search and
  penalized regression.
\newblock \emph{The Annals of Statistics}, \textbf{42}:\penalty0 2526--2556,
  2014.

\bibitem[Chen et~al.(2019)Chen, Drton, and Wang]{chen}
W.~Y. Chen, M.~Drton, and Y.~S. Wang.
\newblock On causal discovery with an equal-variance assumption.
\newblock \emph{Biometrika}, \textbf{106}:\penalty0 973--980, 2019.

\bibitem[Chickering(2003)]{chickering2003}
D.~W. Chickering.
\newblock Optimal structure identification with greedy search.
\newblock \emph{The Journal of Machine Learning Research}, \textbf{3}:\penalty0
  507--554, 2003.

\bibitem[Cormen et~al.(2009)Cormen, Leiserson, Rivest, and
  Stein]{cormen2009introduction}
T.~H. Cormen, C.~E. Leiserson, R.~L. Rivest, and C.~Stein.
\newblock \emph{Introduction to algorithms}.
\newblock MIT Press, 2009.

\bibitem[Erd\"os and R\'enyi(1960)]{erdHos1960evolution}
P.~Erd\"os and A.~R\'enyi.
\newblock On the evolution of random graphs.
\newblock \emph{Publications of the Mathematical Institute of the Hungarian
  Academy of Sciences}, \textbf{5}:\penalty0 17--60, 1960.

\bibitem[Friedman et~al.(2010)Friedman, Hastie, and Tibshirani]{Friedman2010}
J.~Friedman, T.~Hastie, and R.~Tibshirani.
\newblock Regularization paths for generalized linear models via coordinate
  descent.
\newblock \emph{Journal of Statistical Software}, \textbf{33}:\penalty0 1--22,
  2010.

\bibitem[Heckerman et~al.(1995)Heckerman, Geiger, and Chickering]{Heckerman1}
D.~Heckerman, D.~Geiger, and D.~M. Chickering.
\newblock Learning {B}ayesian networks: the combination of knowledge and
  statistical data.
\newblock \emph{Machine Learning}, \textbf{20}:\penalty0 197--243, 1995.

\bibitem[Kalisch and B{\"u}hlmann(2007)]{Kalisch}
M.~Kalisch and P.~B{\"u}hlmann.
\newblock Estimating high-dimensional directed acyclic graphs with the
  {PC}-algorithm.
\newblock \emph{The Journal of Machine Learning Research}, \textbf{8}:\penalty0
  613--636, 2007.

\bibitem[Kalisch et~al.(2012)Kalisch, M{\"a}chler, Colombo, Maathuis, and
  B{\"u}hlmann]{Kalisch2012}
M.~Kalisch, M.~M{\"a}chler, D.~Colombo, M.~H. Maathuis, and P.~B{\"u}hlmann.
\newblock Causal inference using graphical models with the {R} package pcalg.
\newblock \emph{Journal of Statistical Software}, \textbf{47}:\penalty0 1--26,
  2012.

\bibitem[Koller and Friedman(2009)]{koller}
D.~Koller and N.~Friedman.
\newblock \emph{Probabilistic graphical models: principles and techniques}.
\newblock Cambridge, Massachusetts: MIT Press, 2009.

\bibitem[Meinshausen and B\"uhlmann(2006)]{Meinshausen}
N.~Meinshausen and P.~B\"uhlmann.
\newblock High-dimensional graphs and variable selection with the lasso.
\newblock \emph{The Annals of Statistics}, \textbf{34}:\penalty0 1436--1462,
  2006.

\bibitem[Morris(1982)]{Morris}
C.~N. Morris.
\newblock Natural exponential families with quadratic variance functions.
\newblock \emph{The Annals of Statistics}, \textbf{10}:\penalty0 65--80, 1982.

\bibitem[Nandy et~al.(2018)Nandy, Hauser, and Maathuis]{Nandy}
P.~Nandy, A.~Hauser, and M.~H. Maathuis.
\newblock High-dimensional consistency in score-based and hybrid structure
  learning.
\newblock \emph{The Annals of Statistics}, \textbf{46}:\penalty0 3151--3183,
  2018.

\bibitem[Park and Park(2019)]{Park20192}
G.~Park and S.~Park.
\newblock High-dimensional {P}oisson structural equation model learning via
  $\ell_1$-regularized regression.
\newblock \emph{The Journal of Machine Learning Research},
  \textbf{18}:\penalty0 1--41, 2019.

\bibitem[Park and Raskutti(2018)]{Park20191}
G.~Park and G.~Raskutti.
\newblock Learning quadratic variance function {DAG} models via overdispersion
  scoring.
\newblock \emph{The Journal of Machine Learning Research},
  \textbf{18}:\penalty0 1--44, 2018.

\bibitem[Peters and B\"uhlmann(2014)]{buhlmann}
J.~Peters and P.~B\"uhlmann.
\newblock Identifiability of {G}aussian structural equation models with equal
  error variances.
\newblock \emph{Biometrika}, \textbf{101}:\penalty0 219--228, 2014.

\bibitem[Peters et~al.(2014)Peters, Mooij, Janzing, and Sch{\"o}lkopf]{peters}
J.~Peters, J.~M. Mooij, D.~Janzing, and B.~Sch{\"o}lkopf.
\newblock Causal discovery with continuous additive noise models.
\newblock \emph{The Journal of Machine Learning Research},
  \textbf{15}:\penalty0 2009--2053, 2014.

\bibitem[Sachs et~al.(2005)Sachs, Perez, Peer, Lauffenburger, and Nolan]{sach}
K.~Sachs, O.~Perez, D.~Peer, D.~A. Lauffenburger, and G.~P. Nolan.
\newblock Causal protein-signaling networks derived from multiparameter
  single-cell data.
\newblock \emph{Science}, \textbf{308}:\penalty0 523--529, 2005.

\bibitem[Sanford and Moosa(2012)]{sanford}
A.~D. Sanford and I.~A. Moosa.
\newblock A {B}ayesian network structure for operational risk modelling in
  structured finance operations.
\newblock \emph{Journal of the Operational Research Society},
  \textbf{63}:\penalty0 431--444, 2012.

\bibitem[Scutari(2010)]{Scutari}
M.~Scutari.
\newblock Learning {B}ayesian networks with the bnlearn {R} package.
\newblock \emph{Journal of Statistical Software}, \textbf{35}:\penalty0 1--22,
  2010.

\bibitem[Shimizu et~al.(2006)Shimizu, Hoyer, Hyv{\"a}rinen, and
  Kerminen]{shimizu2006}
S.~Shimizu, P.~O. Hoyer, A.~Hyv{\"a}rinen, and A.~J. Kerminen.
\newblock A linear non-{G}aussian acyclic model for causal discovery.
\newblock \emph{The Journal of Machine Learning Research}, \textbf{7}:\penalty0
  2003--2030, 2006.

\bibitem[Shimizu et~al.(2011)Shimizu, Inazumi, Sogawa, Hyv{\"a}rinen, Kawahara,
  Washio, Hoyer, and Bollen]{shimizu2011}
S.~Shimizu, T.~Inazumi, Y.~Sogawa, A.~Hyv{\"a}rinen, Y.~Kawahara, T.~Washio,
  P.~O. Hoyer, and K.~Bollen.
\newblock {DirectLiNGAM}: a direct method for learning a linear non-{G}aussian
  structural equation model.
\newblock \emph{The Journal of Machine Learning Research},
  \textbf{12}:\penalty0 1225--1248, 2011.

\bibitem[Spirtes et~al.(2000)Spirtes, Glymour, and Scheines]{Spirtes2000}
P.~Spirtes, C.~N. Glymour, and R.~Scheines.
\newblock \emph{Causation, {P}rediction, and {S}earch}.
\newblock Cambridge, Massachusetts: MIT Press, 2000.

\bibitem[Sun et~al.(2013)Sun, Wang, and Fang]{SunWW2013}
W.~W. Sun, J.~H. Wang, and Y.~X. Fang.
\newblock Consistent selection of tuning parameters via variable selection
  stability.
\newblock \emph{The Journal of Machine Learning Research},
  \textbf{14}:\penalty0 3419--3440, 2013.

\bibitem[Tsamardinos et~al.(2006)Tsamardinos, Brown, and Aliferis]{Tsama}
I.~Tsamardinos, L.~E. Brown, and C.~F. Aliferis.
\newblock The max-min hill-climbing {B}ayesian network structure learning
  algorithm.
\newblock \emph{Machine Learning}, \textbf{65}:\penalty0 31--78, 2006.

\bibitem[Wang and Drton(2020)]{WangYS2020}
Y.~S. Wang and M.~Drton.
\newblock High-dimensional causal discovery under non-{G}aussianity.
\newblock \emph{Biometrika}, \textbf{107}:\penalty0 41--59, 2020.

\bibitem[Yang et~al.(2015)Yang, Ravikumar, Allen, and Liu]{yang2015}
E.~Yang, P.~Ravikumar, G.~I. Allen, and Z.~D. Liu.
\newblock Graphical models via univariate exponential family distributions.
\newblock \emph{The Journal of Machine Learning Research},
  \textbf{16}:\penalty0 3813--3847, 2015.

\bibitem[Yuan et~al.(2019)Yuan, Shen, Pan, and Wang]{Yuan2019}
Y.~P. Yuan, X.~T. Shen, W.~Pan, and Z.~Z. Wang.
\newblock Constrained likelihood for reconstructing a directed acyclic
  {G}aussian graph.
\newblock \emph{Biometrika}, \textbf{106}:\penalty0 109--125, 2019.

\bibitem[Zheng et~al.(2018)Zheng, Aragam, Ravikumar, and Xing]{Zheng}
X.~Zheng, B.~Aragam, P.~K. Ravikumar, and E.~P. Xing.
\newblock {DAG}s with {NO TEARS}: {C}ontinuous optimization for structure
  learning.
\newblock \emph{In Advances in Neural Information Processing Systems (NIPS)},
  pages 9472--9483, 2018.

\bibitem[Zhu et~al.(2020)Zhu, Ng, and Chen]{Zhu}
S.~Y. Zhu, I.~Ng, and Z.~T. Chen.
\newblock Causal discovery with reinforcement learning.
\newblock \emph{International Conference on Learning Representations (ICLR)},
  2020.

\end{thebibliography}
\bibliographystyle{plainnat}

\end{document}